\documentclass[]{arxiv/fairmeta}
\usepackage{amsmath,amsfonts,bm}

\def\eqref#1{equation~\ref{#1}}
\def\1{\bm{1}}

\DeclareMathAlphabet{\mathsfit}{\encodingdefault}{\sfdefault}{m}{sl}
\SetMathAlphabet{\mathsfit}{bold}{\encodingdefault}{\sfdefault}{bx}{n}

\usepackage{hyperref}
\usepackage{url}
\usepackage[most]{tcolorbox}
\usepackage{xcolor}
\usepackage{colortbl}
\usepackage{booktabs}
\usepackage{listings}
\usepackage{fancyvrb}
\usepackage{xcolor}

\usepackage{enumitem}
\usepackage{graphicx}
\usepackage{cleveref}

\usepackage{algorithm}
\usepackage{algpseudocode}   % cleaner pseudocode environment (ICML does not support for some reason)
\usepackage{wrapfig}
\usepackage{tcolorbox}
\usepackage{wrapfig}
\usepackage{multirow}
\usepackage{caption}

\tcbuselibrary{skins, breakable}

\definecolor{darkred}{RGB}{139,0,0}
\definecolor{darkgreen}{RGB}{0,139,0}
\definecolor{darkblue}{RGB}{0,0,139}
\definecolor{metablue}{HTML}{0064E0}

\newcommand{\COne}{\textcolor{darkred}{\textbf{C1}}}
\newcommand{\CTwo}{\textcolor{darkred}{\textbf{C2}}}

\usepackage{tcolorbox}
\tcbuselibrary{listings,breakable,skins}
\usepackage{listings}

\lstdefinestyle{colcode}{
  basicstyle=\ttfamily\footnotesize,
  columns=fullflexible,
  breaklines=true,
  breakatwhitespace=true,
  keepspaces=true,
  showstringspaces=false
}

\usepackage{listings}
\title{
Confucius Code Agent: Scalable Agent Scaffolding for Real-World Codebases
}

\author[1,*]{\textcolor{metablue}{Sherman Wong}}
\author[2,*]{\textcolor{metablue}{Zhenting Qi}}
\author[1,*]{\textcolor{metablue}{Zhaodong Wang}}
\author[1,*]{\textcolor{metablue}{Nathan Hu}}
\author[1]{\\Samuel Lin}
\author[1]{Jun Ge}
\author[1]{Erwin Gao}
\author[1]{Wenlin Chen}
\author[2]{Yilun Du}
\author[1,2]{Minlan Yu}
\author[1]{Ying Zhang}
\affiliation[1]{Meta}
\affiliation[2]{Harvard}

\contribution[*]{\textcolor{metablue}{\textbf{Core Contributors}}}

\abstract{
Real-world software engineering tasks require coding agents that can operate on massive repositories, sustain long-horizon sessions, and reliably coordinate complex toolchains at test time. Existing research-grade coding agents offer transparency but struggle when scaled to heavier, production-level workloads, while production-grade systems achieve strong practical performance but provide limited extensibility, interpretability, and controllability.
We introduce the \textbf{Confucius Code Agent (CCA)}, a software engineering agent that can operate at large-scale codebases. CCA is built on top of the \textbf{Confucius SDK}, an agent development platform structured around three complementary perspectives: Agent Experience (AX), User Experience (UX), and Developer Experience (DX). The SDK supports a unified orchestrator with advanced context management for long-context reasoning, a persistent note-taking system for cross-session continual learning, and a modular extension system for reliable tool use.
In addition, we introduce a \textbf{meta-agent} that automates the construction, evaluation, and refinement of agents through a build-test-improve cycle, enabling rapid agent development on new tasks and tool stacks.
Instantiated on the Confucius SDK using the meta-agent, CCA demonstrates strong performance on real-world software engineering tasks. On SWE-Bench-Pro, CCA achieves a Resolve@1 of \textbf{59\%}, exceeding prior research baselines as well as commercial results, under identical repositories, model backends, and tool access.
}

\date{\today}
\correspondence{Zhenting Qi at \email{zhentingqi@g.harvard.edu}}
\github{https://github.com/facebookresearch/cca-swebench}

\begin{document}

\maketitle

\begin{figure*}[h]
    \centering
    \includegraphics[width=\textwidth]{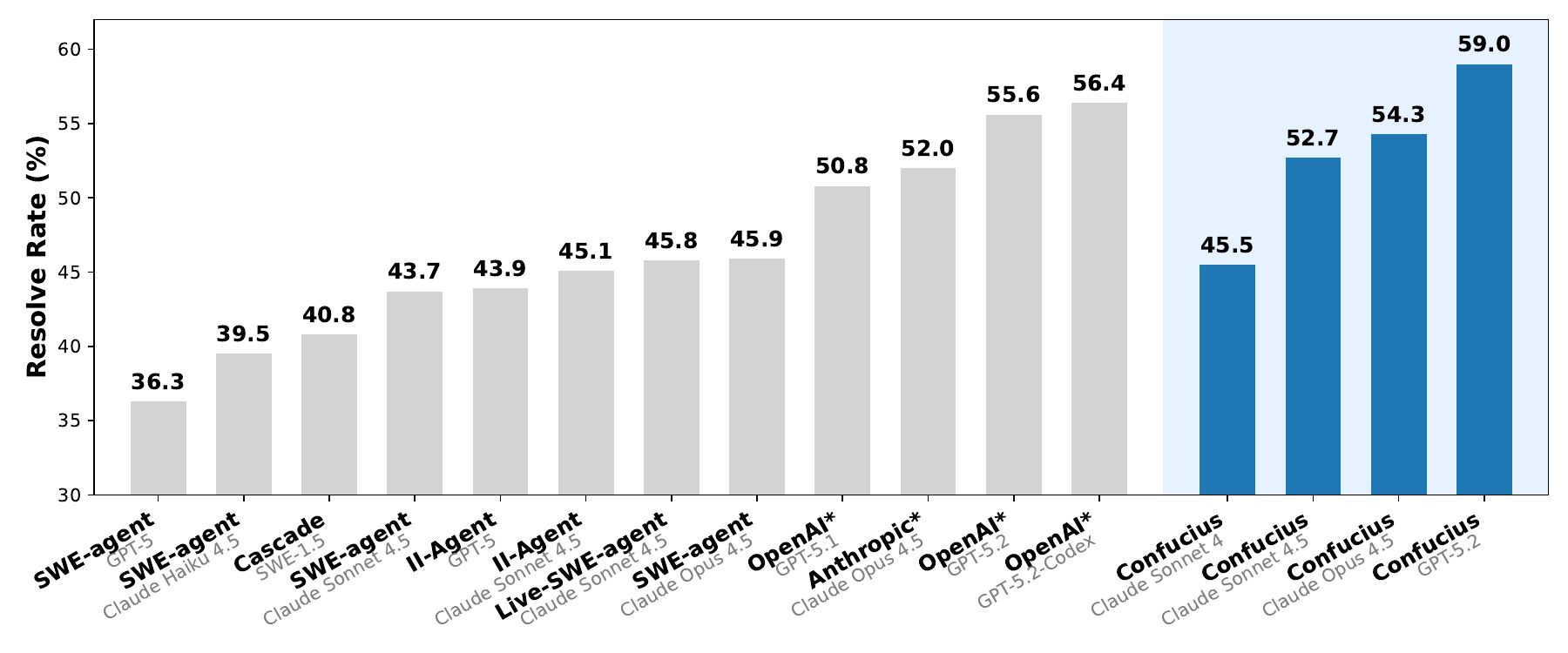}
    \caption{\textbf{Performance comparison on SWE-Bench-Pro benchmark.} (* reported from Anthropic's Claude Opus 4.5 system card / OpenAI's GPT-5.2-Codex system card.)}
    \label{fig:teaser}
\end{figure*}

\section{Introduction}

Software engineering has rapidly emerged as a frontier application area for large language models (LLMs).  
As models have grown more capable, they have progressed from simple program synthesis \citep{austin2021program}, to automatic code completion \citep{humaneval}, to general-purpose code generation \citep{li2022competition, lai2023ds}, to understanding code execution \citep{gu2024cruxeval}, and competition-level programming \citep{jain2024livecodebench}.  
Most recently, LLMs have demonstrated strong software engineering ability to tackle real-world issue resolution in open-source repositories \citep{swebench, sweagent, xia2025live, satoriswe}.  
To support such capabilities, more sophisticated agentic frameworks such as OpenHands \citep{openhands} scaffold LLMs with tools for search, code editing, and command execution, while agentless prompting-based approaches \citep{xia2024agentless} have shown that carefully structured prompts alone can also perform well on multi-step software engineering tasks. 

While model capabilities continue to improve, success in real-world software engineering depends not only on the underlying LLM, but also on the \emph{agent scaffold}: the orchestration, memory structures, and tool abstractions surrounding the model. Empirically, even when the same backbone model is used, different scaffolding strategies can lead to large performance disparities \citep{xia2025live}, suggesting that the design of the agent’s cognitive and operational environment is a fundamental research dimension. However, existing coding agents often rely on flat interaction histories, heuristic prompt engineering, or tightly coupled tool pipelines, which are difficult to scale to the long-horizon, multi-file, multi-step workflows characteristic of enterprise-level software engineering.
This gap is most clearly manifested in two core \textcolor{darkred}{\textbf{challenges}}:
\begin{itemize}[leftmargin=*]
    \item \textbf{\COne: Long-context reasoning.} Agents must efficiently localize relevant code within massive repositories and perform multi-hop reasoning across dispersed modules, long tool traces, and deep execution histories.
    \item \textbf{\CTwo: Long-term memory.} Agents should accumulate persistent knowledge across tasks and sessions, capturing reusable patterns, failure modes, and invariants, rather than repeatedly rediscovering information or reproducing past mistakes.
\end{itemize}
These challenges highlight that scalability in agentic software engineering requires more than longer context windows or larger models: it requires a principled approach to how agents structure, maintain, and interact with external information.

We argue that addressing these challenges requires a broader, system-level design perspective. In particular, we decompose agentic interaction into three complementary axes: \textbf{Agent Experience (AX)}, \textbf{User Experience (UX)}, and \textbf{Developer Experience (DX)}.  
AX concerns the agent’s internal cognitive workspace: how information is distilled, organized, and presented to the LLM for stable reasoning.  
UX concerns the transparency, controllability, and interpretability required for human users to understand the agent’s behavior.  
DX concerns observability, evaluation, and modularity for researchers and practitioners developing and improving agent systems.  
Most existing frameworks conflate these axes. For example, passing human-oriented logs directly into the agent’s prompt, thereby degrading AX, limiting UX, and restricting DX. Treating AX, UX, and DX as first-class and distinct design principles provides a foundation for scalable, analyzable, and reproducible agent behavior.

We first introduce the \textbf{Confucius SDK} (\Cref{fig:teaser}), an agent development platform explicitly structured around AX, UX, and DX. 
On this platform, we instantiate the \textbf{Confucius Code Agent (CCA)}, a concrete agent tailored to large-scale software engineering. 
We introduce four key mechanisms, each served for AX, UX, and/or DX, and instantiated in CCA to address the two \textcolor{darkred}{\textbf{challenges}} above:
\begin{enumerate}[leftmargin=*]
    \item \textbf{Context management (AX; \COne).}  
    A hierarchical working memory coupled with context compression enables the agent to retain essential state while supporting long-horizon reasoning without exceeding context limits.

    \item \textbf{Note-taking (AX, UX; \CTwo).}  
    A dedicated note-taking agent distills trajectories into persistent, hierarchical Markdown notes, including hindsight notes that capture failure modes, thereby supporting both durable knowledge for the agent (AX) and interpretable artifacts for humans (UX).

    \item \textbf{Extensions (AX, DX; \COne).}  
    Modular extensions define tool-use behavior, parsing, prompt shaping, and interaction policies through typed callbacks. This separation improves agent control and reasoning stability (AX) while providing observability and composability for developers (DX). CCA binds together coding-specific extensions such as file search, file editing, and CLI tools. 

    \item \textbf{Meta-agent (DX).}  
    A meta-agent automates a build-test-improve loop that synthesizes, evaluates, and refines agent configurations, enabling rapid agent development and adaptation to new environments.
\end{enumerate}

% Throughout this paper, we distinguish between the Confucius SDK and the CCA. The Confucius SDK is a general-purpose agent development platform that provides reusable abstractions for orchestration, memory, and tool use, structured around AX, UX, and DX. The CCA is a single concrete instantiation built on top of this SDK, configured specifically for large-scale software engineering tasks. Our empirical results are obtained using CCA, while the design principles and mechanisms we describe apply more broadly to agents constructed using the Confucius SDK.

To evaluate the mechanisms introduced by Confucius SDK and CCA, we conduct experiments on SWE-Bench-Verified \citep{swebench} and SWE-Bench-Pro \citep{sbp}.
CCA achieves strong performance compared with prior coding agents, demonstrating how principled scaffolding can substantially amplify the effectiveness of the same underlying LLM.
We also evaluate CCA on a custom PyTorch-Bench targeting debugging workflows on larger-scale codebases. Ablations isolate the contributions of each crucial mechanism, and analyses categorize remaining failure cases. In summary, our contributions are:
\begin{itemize}[leftmargin=*]
    \item We present the \textbf{Confucius Code Agent (CCA)}, a coding agent designed for large-scale software engineering.
    \item We present the \textbf{Confucius SDK}, a principled AX/UX/DX-balanced agent development platform with advanced context management, extensions, and long-term memory.
    \item We provide evaluations across multiple benchmarks, supported by ablations and case studies. Notably, on SWE-Bench-Pro, CCA reaches a Resolve@1 performance of \textbf{59\%}, exceeding prior research baselines and  commercial results, under identical repositories, model backend, and tool access.
    \item We empirically show that \emph{agent scaffolding}, not just model capability, is also a primary determinant of agent performance, with appropriate orchestration and memory structures outperforming stronger models.
\end{itemize}

\begin{figure*}[t]
    \centering
    \includegraphics[width=\textwidth]{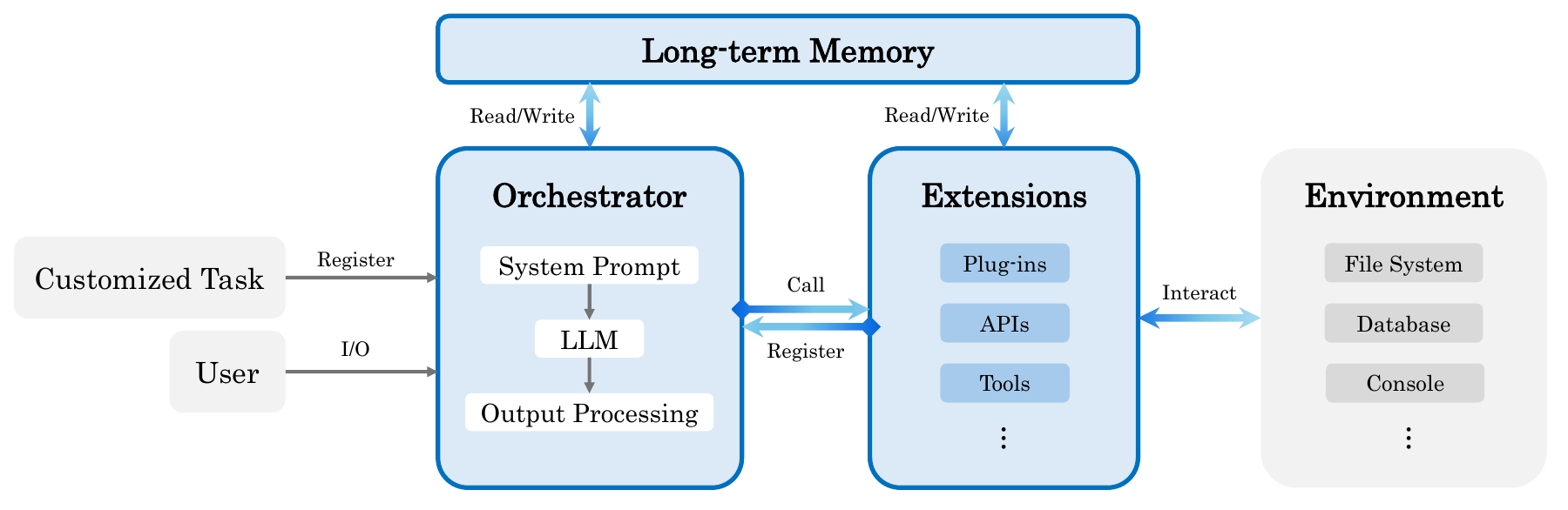}
    \caption{\textbf{Confucius SDK overview.}  
    The SDK unifies an orchestrator for iterative reasoning and action execution, long-term memory for continual learning, and modular extensions for tool use and interacting with the external environment.}
    \label{fig:teaser}
\end{figure*}

\section{Method}
\label{sec.method}

In this section, we describe both the algorithmic design of agentic reasoning and the system design of the Confucius SDK that supports it. We first introduce the overarching design philosophy (AX, UX, DX) (\Cref{sec.3x}), then separate the presentation into (i) the SDK abstractions that realize these design choices in a modular and extensible manner and how they are instantiated on CCA (\Cref{sec.feat}) and (ii) the core agentic algorithms that govern how agents reason and improve over time (\Cref{sec.agentic_algo}). 

\subsection{Design Philosophy: AX, UX, and DX}
\label{sec.3x}

% \begin{wrapfigure}{r}{0.45\columnwidth}
%   \centering
%   \includegraphics[width=\linewidth]{figs/3X.pdf}
%   \caption{Wrapped figure}
% \end{wrapfigure}

Most agent frameworks optimize for a single audience (users, agents, or developers). In contrast, the Confucius SDK adopts a three-axis philosophy that treats \emph{Agent Experience (AX)}, \emph{User Experience (UX)}, and \emph{Developer Experience (DX)} as distinct yet coupled design goals.
AX concerns the agent’s cognitive workspace: what context the model receives, how it is structured, and how it invokes tools. It should be concise and structured to avoid distraction or bias.
UX governs how humans observe and interact with the system, prioritizing transparency via readable logs, traces, and artifact previews.
DX supports building and improving agents through observability and control over prompts, tools, and memory, with visibility into both AX and UX.

Many systems conflate AX and UX by feeding human-facing traces directly to the model, causing context bloat, spurious anchors, and brittle debugging. CCA instead separates channels: users get rich instrumented traces, the agent gets compressed structured inputs, and developers can inspect and modify both.

Below shows a concrete example of the distinction between AX and UX: \\

\noindent
\begin{tcolorbox}[
  width=\columnwidth,
  colback=white,
  colframe=black!25,
  boxrule=0.4pt,
  arc=2pt,
  left=4pt,right=4pt,top=4pt,bottom=4pt,
  breakable
]
\textbf{For UX (Users See):}
\vspace{2pt}
\begin{lstlisting}[style=colcode]
Creating file at config.py
File created successfully at config.py
Here is the diff:
+ PORT=8080
+ DEBUG=true
+ MAX_CONNECTIONS=100
\end{lstlisting}

\vspace{0.4em}
\hrule
\vspace{0.4em}

\textbf{For AX (Agent Sees):}
\vspace{2pt}
\begin{lstlisting}[style=colcode]
Human: [previous user message]
AI: <file_edit type="create" file_path="config.py">...</file_edit>
Human: <result>File created successfully</result>
\end{lstlisting}
\end{tcolorbox}

In this case, users are presented with rich, streaming updates, whereas the agent receives only a compressed summary of the outcome stored in the memory manager, without the lengthy file diff message.

\subsection{Confucius SDK Features \& Instantiations on CCA}
\label{sec.feat}

\subsubsection{Context Management}
\label{sec.context_management}

Running agents on large-scale repositories quickly stresses even long-context LLMs: long debugging sessions, multi-file refactors, and nested tool calls all contribute to unbounded conversation growth. In many existing coding agent frameworks, agents either accumulate a single flat history (risking hard context limits and ``forgotten'' early decisions) or rely on naive truncation and ad-hoc retrieval, which can silently drop important information and are difficult to tune for different workloads. 
The Confucius SDK addresses this by providing an explicit \emph{agent context management} layer that combines hierarchical working memory with adaptive context compression.

\begin{figure*}[t]
    \centering
    \includegraphics[width=\textwidth]{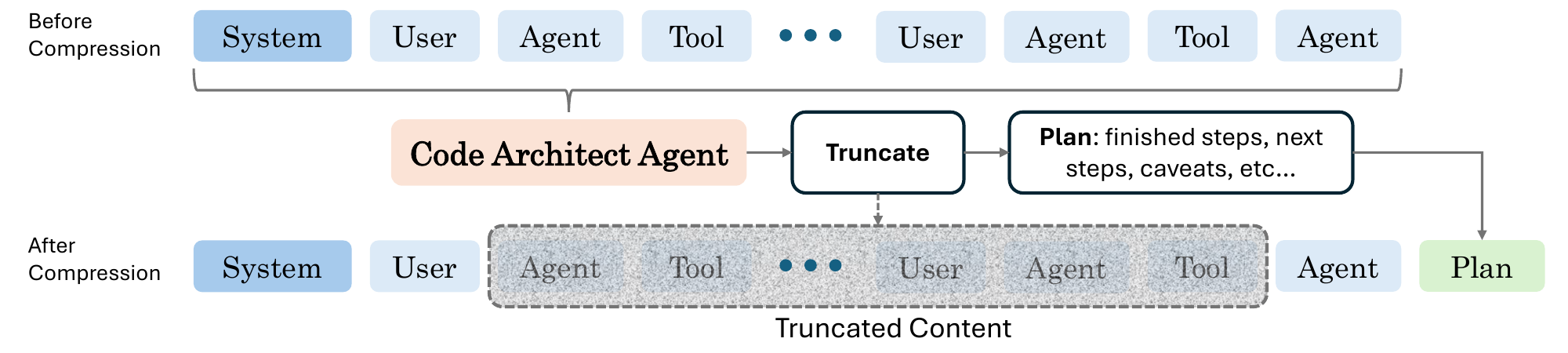}
    \caption{\textbf{Context compression overview.}
    When the context window approaches configurable thresholds, the \textit{Architect} agent summarizes earlier turns into a structured plan containing goals, decisions, errors, and open TODOs. These compressed summaries replace original large spans of history while preserving a short window of recent interactions, enabling the agent to sustain multi-step reasoning over long trajectories without exceeding context limits.}
    \label{fig:mem_compress}
\end{figure*}

At the SDK level, each instantiated agent is backed by a file-system-based \textbf{hierarchical working memory} with configurable visibility scopes (e.g., session, entry, runnable). 
The memory is organized as a tree-structured namespace where internal nodes represent semantic groupings and leaf nodes store Markdown documents annotated with metadata tags. Agents interact with this memory through a set of primitive operations, including search, read, write, edit, and delete, which are exposed as callable tools during generation.
Below is an example of a constructed hierarchical memory of CCA on a SWE-Bench-Pro instance. 
The agent maintains this hierarchy throughout execution so that when context is pruned, important insights and intermediate artifacts can be stored and later retrieved efficiently:

\begin{tcolorbox}[colback=white,  % background color
                   colframe=black,    % frame color
                   coltext=metablue,      % text color
                   breakable,
                   fontupper=\scriptsize\fontfamily{pcr}\selectfont]  % font style and size
\begin{verbatim}
+-- instance_qutebrowser__qutebrowser-c09e1439...
    +-- hierarchical_memory_3a7488c6-bf8c-11f0-...
        +-- qutebrowser_process_cleanup
            |-- analysis.md
            |-- implementation_summary.md
            +-- todo.md
\end{verbatim}
\end{tcolorbox}

On top of this hierarchy, Confucius SDK integrates an adaptive \textbf{context compression} mechanism (\Cref{fig:mem_compress}) driven by an \emph{Architect Agent}. When the context length approaches configurable thresholds, the Architect Agent is invoked in a separate call to analyze the conversation history and construct a structured summary that explicitly preserves key information categories (e.g., task goals, decisions made, open TODOs, and critical error traces). The system then replaces marked historical messages with this compressed summary while maintaining a rolling window of recent messages in their original form. The summary is inserted as a new message, and all future turns will see both the compact summary and the recent raw history. 

This context management design provides two key benefits.
First, structured summarization triggered only when needed preserves semantically important information and maintains access to long reasoning chains, avoiding the brittleness of fixed-window truncation or simple retrieval.
Second, the hierarchical memory stores and refines key insights throughout execution, complementing the summaries and ensuring that important state persists even as the raw history is compressed.
In CCA, these mechanisms are essential for handling long-running software engineering sessions on industrial-scale codebases, improving performance on long-context coding tasks without requiring changes to the underlying orchestrator or extensions. 
Similar context engineering techniques are also reported in recent production-level LLMs \citep{anthropic2025contextengineering, openai2025sessionmemory}.

\subsubsection{Note-Taking}
\label{sec.note_taking}

Flat chat logs are not an ideal representation for long-term memory: they are verbose and difficult to reuse without manually rereading entire transcripts. In typical frameworks, any cross-session ``memory'' is either absent or implemented via coarse-grained embeddings over whole turns, which tends to miss important structure such as architectures, design decisions, and failure modes. To support agents that improve over time and can pick up long-running projects where they left off, the Confucius SDK includes an explicit \emph{note-taking} mechanism that memorizes interaction traces as structured persistent knowledge.

At the SDK level, every interaction session is logged into a structured ``trajectory'', including user messages, tool invocations, LLM outputs, and system events. A dedicated \textbf{note-taking agent} can distill these trajectories into compact notes without affecting the online latency of the primary agent. Persistent notes are stored as Markdown files in a file-system-like tree: each session has an associated directory, under which the note-taking agent can create paths such as \texttt{research/findings.md} and  \texttt{solutions/bug\_fix.md}. Leaves in this hierarchy are Markdown documents with lightweight tags, maintained as typed memory nodes. The SDK exposes structured tools to search, read, write, edit, delete, and import these nodes, so notes can be programmatically updated and reused across sessions. Examples of taken notes are shown in Appendix~\ref{app:notes}.

A distinctive aspect of the Confucius SDK’s note-taking is its emphasis on \emph{hindsight notes} for failures. The note-taking layer encourages agents to record not only successful solutions but also compilation errors, runtime exceptions, and unproductive strategies, together with eventual resolutions or reasons for abandonment. Over time, this yields a corpus of failure cases indexed by error messages, stack traces, and affected components. When a similar failure appears in a future session, an agent can retrieve the corresponding hindsight note and immediately surface known fixes or workarounds, rather than rediscovering them from scratch. In CCA, these mechanisms turn day-to-day usage on large codebases into a steadily growing, human-readable body of durable knowledge that improves continuity across sessions and reduces repeated “thrashing” on recurring issues.

\subsubsection{Extensions}
\label{sec.extensions}

The Confucius SDK implements output parsing, tool invocation, and side-effect management via \emph{extensions}: modular components that attach to the orchestrator and run at every execution step. Whereas many frameworks hard-code these behaviors in ad-hoc logic or model-specific prompting, Confucius exposes them as first-class, composable modules that are easier to reuse, audit, and adapt.

Each extension is a typed configuration object that registers ordered callbacks (e.g., \texttt{on\_input\_messages}, \texttt{on\_llm\_output}) executed within the orchestrator loop. Callbacks share a run context providing I/O, session state, memory, and artifacts, enabling extensions to shape prompts, interpret structured outputs (XML tags or native tool calls), and inject or filter messages while maintaining local state.

Conceptually, extensions cover perception, reasoning, and action: they (i) parse and validate model outputs into structured actions, (ii) rewrite or annotate inputs before LLM calls, and (iii) execute tools (e.g., shell, file edits) and summarize results into memory or artifacts. By routing tool use and prompt shaping through extensions, Confucius cleanly separates orchestration from capabilities. CCA is thus an orchestrator paired with a particular extension bundle; our ablations (\Cref{tab:context_mgmt}) vary only enabled extensions and configurations while holding the loop fixed, so improvements to extensions transfer across all Confucius-based agents.

\subsection{Agentic Algorithms: Orchestration and Automatic Agent Development}
\label{sec.agentic_algo}

\subsubsection{The Confucius Orchestrator}
\label{sec.orchestrator}

\begin{wrapfigure}{r}{0.6\textwidth}
\vspace{-2em}
\begin{tcolorbox}[
    enhanced,
    colback=white,
    colframe=black!40,
    sharp corners,
    boxrule=0.5pt,
    left=4pt,
    right=4pt,
    top=4pt,
    bottom=4pt
]

\hrule height 0.8pt \vspace{0.3em}

\textbf{Algorithm 1: Confucius Orchestrator Loop}

\hrule height 0.4pt \vspace{0.3em}

\begin{algorithmic}[1]
    \State \textbf{Initialize} session context, memory, extensions
    \While{iteration $< \texttt{max\_iters}$}
        \State \textbf{Invoke} LLM with system prompt + memory
        \State \textbf{Parse} LLM output into actions
        \ForAll{actions $a$}
            \State \textbf{Route} $a$ to its extension
            \State \textbf{Execute} extension; update memory
            \If{extension signals continuation}
                \State add observations (results, error, etc.) to memory
                \State \textbf{continue}
            \EndIf
        \EndFor
        \State \textbf{Check} for completion; \textbf{break} if done
    \EndWhile
    \State \textbf{return} final output and artifacts
\end{algorithmic}

\hrule height 0.4pt

\end{tcolorbox}
\vspace{-2em}
\end{wrapfigure}

% \begin{algorithm}[tb]
%    \caption{Confucius Orchestrator Loop}
%    \label{alg:confucius_orchestrator}
% \begin{algorithmic}[1]
%    \STATE \textbf{Initialize} session context, memory, and extensions.
%    \WHILE{iteration $< \texttt{max\_iters}$}
%       \STATE \textbf{Invoke} LLM with system prompt and memory.
%       \STATE \textbf{Parse} LLM output into actions.
%       \FORALL{actions $a$}
%          \STATE \textbf{Route} $a$ to its extension.
%          \STATE \textbf{Execute} extension and update memory.
%          \IF{extension signals continuation}
%             \STATE Add observations (e.g. errors) to memory.
%             \STATE \textbf{continue}
%          \ENDIF
%       \ENDFOR
%       \STATE \textbf{Check} for completion; \textbf{break} if done.
%    \ENDWHILE
%    \STATE \textbf{return} final output and artifacts.
% \end{algorithmic}
% \end{algorithm}

The Confucius Orchestrator is a minimal, extensible loop that repeatedly calls the LLM, interprets its outputs, and coordinates tool execution. The same orchestrator is reused across all Confucius SDK agents; CCA is one configuration of this loop with coding-oriented extensions enabled.

\paragraph{Output Processing.}
The orchestrator supports two LLM interfaces: (1) native tool-use models emit structured tool calls that are routed to extension handlers; (2) other models emit XML-style tags (e.g., \texttt{<bash>...</bash>}), which are parsed into the same action representation. This preserves broad model compatibility while leveraging native tool-use when available.

\paragraph{Iteration Control.}
Execution is bounded by a maximum iteration cap, but termination is typically agent-driven. Each iteration parses the LLM output into actions; if no further actions are produced, the run ends. Extensions can also force continuation (e.g., a bash execution returns an interrupt with command output), causing the orchestrator to re-invoke the LLM with updated context. Together, these mechanisms support multi-step reasoning and tool use within safe bounds.

\subsubsection{Meta-agent: A Build-Test-Improve Loop}
\label{sec.meta_agent}

A recurring limitation of existing agent frameworks is that agent behavior is largely \emph{static}: humans hand-design prompts, tool wiring, and guardrails, then periodically revise them by trial and error. This is labor-intensive and does not scale with growing tool ecosystems. Moreover, we find that naive implementations of file-editing or command-line tools, even when they are functionally correct, often underperform because the surrounding prompts and error-handling conventions are not tuned to realistic workloads. We address this by introducing a \emph{Meta Agent}, an agent that automatically builds and refines other agents through an explicit \emph{build-test-improve} loop, turning agent design itself into an agentic, evaluation-driven automatic process.
More implementation details of Meta Agent can be found at Appendix \ref{sec.meta-agent-details}.

This Meta-agent capability lives at the Confucius SDK level but directly benefits CCA. The proposed CCA is itself the outcome of the Meta-agent's build-improve-test loop: we start from a high-level description of a repository-level software engineering assistant, let the Meta-agent synthesize the orchestrator configuration, tool wiring, and prompts, and then repeatedly refine them against a production-grade test set until performance stabilizes. 
% The resulting agent exhibits more reliable tool selection and recovery behaviors than our initial hand-written designs, and these improvements are reflected in the tool-use ablations reported in \Cref{sec:meta-agent-tool-use}.
% At the same time, the same Meta-agent interface allows users to rapidly spin up use-case-specific agents using this automated iteration loop.
% extending this process to a broader family of specialized agents is an active direction for future work.

\section{Experiments}

\subsection{Setup}

\paragraph{Models and agent scaffold.}
We use Claude 4 Sonnet, Claude 4.5 Sonnet, Claude 4.5 Opus, and GPT-5.2 as the primary backbone LLMs to ensure comparability with published baselines. We use SWE-Agent \citep{sweagent} as the baseline scaffold. Our CCA agent replaces the SWE-Agent stack with the Confucius Code Agent.
We also report results from the Live-SWE-Agent \citep{xia2025live} as baseline, while keeping the tool environment and repository setup identical. 

\paragraph{Benchmark.}
For main results, we evaluate CCA on the SWE-Bench-Pro \citep{sbp} public split consisting of 731 tasks, following the identical environment configuration and infrastructure used by the SWE-Agent baseline \citep{sweagent}. We also report results from the SWE-Bench-Verified \citep{swebench} consisting of 500 tasks to compare with existing open-sourced coding agents, including SWE-Agent and OpenHands \citep{openhands}.

\paragraph{Metrics.}
We follow the official SWE-Bench-Pro Resolve Rate metric, defined as the percentage of tasks for which the agent’s proposed patch successfully passes all repository-provided tests without human intervention.\footnote{\url{https://scale.com/leaderboard/swe_bench_pro_public}} Each trial is repeated with different random seeds for trajectory sampling to account for randomness in tool invocation and LLM responses. We report mean Resolve Rate (Pass@1) across three runs.

\subsection{Main Results on SWE-Bench-Pro}

\begin{table}[h]
\centering
\begin{tabular}{l l c}
\toprule
\textbf{Backbone Model} & \textbf{Scaffold} & \textbf{Resolve Rate (Pass@1)} \\
\midrule

\multirow{2}{*}{Claude 4 Sonnet}
    & SWE-Agent \citep{sweagent} & 42.7 \\
    & CCA           & 45.5 \\
\midrule

\multirow{3}{*}{Claude 4.5 Sonnet}
    & SWE-Agent  & 43.6 \\
    & Live-SWE-Agent \citep{xia2025live}       & 45.8 \\
    & CCA            & 52.7 \\
\midrule

\multirow{2}{*}{Claude 4.5 Opus}
    & Anthropic*  & 52.0 \\
    & CCA            & \textbf{54.3} \\
\midrule

\multirow{2}{*}{GPT-5.2}
    & OpenAI* & 56.0 \\
    & CCA     & \textbf{59.0} \\
\bottomrule
\end{tabular}
\caption{SWE-Bench-Pro public split comparison across scaffolds and backbone models.
All methods share identical environments; improvements arise solely from the agent scaffolds.
(* Anthropic's proprietary scaffold, from Claude Opus 4.5 system card; * OpenAI's proprietary scaffold, from GPT-5.2 report.)}
\label{tab:swepro}
\end{table}

\begin{table}[t]
\centering
\begin{tabular}{l l l c}
\toprule
\textbf{Backbone Model} & \textbf{Context Management} & \textbf{Tool Use} & \textbf{Resolve Rate (Pass@1)} \\
\midrule

\multirow{2}{*}{Claude 4 Sonnet}
    & No  & advanced & 42.0 \\
    & Yes & advanced & 48.6 \\
\midrule

\multirow{3}{*}{Claude 4.5 Sonnet}
    & No  & simple   & 44.0 \\
    & No  & advanced & 51.0 \\
    & Yes & advanced & \textbf{51.6} \\
\bottomrule
\end{tabular}
\caption{Ablation on context management (hierarchichal working memory \& context compression) and tool-use sophistication. Results are obtained from evaluating CCA on a 100-example subset of the SWE-Bench-Pro public set.}
\label{tab:context_mgmt}
\end{table}

\Cref{tab:swepro} summarizes our main results on the SWE-Bench-Pro public split. Under identical environment and tool conditions, CCA consistently surpasses the SWE-Agent baseline across settings with different backbone models. 
With Claude 4 Sonnet, CCA reaches Resolve@1 at \textbf{45.5\%}. With Claude 4.5 Sonnet, CCA reaches \textbf{52.7\%}, largely surpassing the best research-grade coding agent, Live-SWE-Agent, at 45.8\%.
With Claude 4.5 Opus, CCA achieves \textbf{54.3\%}, achieving higher performance than results reported by Anthropic. 
And with GPT-5.2, we achieve \textbf{59.0\%}, surpassing the official result reported by OpenAI and setting new leading performance on SWE-Bench-Pro.
These improvements arise purely from stronger agentic scaffolding, i.e. enhanced orchestration, context management, and tool-use extensions, rather than differences in backbone models or evaluation setups.
More broadly, these results underscore the central role of scaffolding: even a weaker model equipped with a strong agent scaffold (Claude 4.5 Sonnet + CCA at \textbf{52.7\%}) can outperform a stronger model (Claude 4.5 Opus + Anthropic's proprietary scaffold at \textbf{52.0\%}). 

Because Claude Code (CC) does not expose a programmatic tool interface compatible with containerized evaluation environments such as SWE-rex \citep{SWEReX2025}, we cannot compare CCA with CC's results on SWE-Bench-Pro. Instead, to provide a qualitative comparison, we constructed a small curated benchmark (a mini PyTorch-Bench) and executed CC solutions using the Claude Code CLI directly on a host machine where CC is installed w/o a Docker-based runtime, as seen in Appendix~\ref{sec:case_studies}. These complementary experiments highlight behavioral differences between CC and CCA in realistic debugging and development tasks, but they are not directly comparable to SWE-Bench-Pro due to differences in execution environments.

\subsection{Meta-Agent Learned Tool-Use}
\label{sec:meta-agent-tool-use}

CCA’s tool-use behavior is not purely hand-engineered; instead, it is \emph{learned} through the Meta-agent, which automatically refines how the agent invokes tools such as file editors and command-line utilities. 
To measure the contribution of this learned tool-use stack, we perform an ablation that disables these Meta-agent–derived tools and instead reverts CCA to a simpler,    ``naive'' tool-use pattern similar to traditional SWE-Agent-like scaffolds with simple file editing and command-line operations only.
\Cref{tab:context_mgmt} reports the results of this ablation alongside a separate ablation on context management. Experiments are conducted on a 100-example subset of the SWE-Bench-Pro public set. As shown in the Claude 4.5 Sonnet rows, removing the learned tool-use features leads to a large decline in Resolve@1, even when the context management method is held constant. This confirms that tool-use conventions learned by the Meta-agent are a major driver of CCA’s performance, independent of (and complementary to) hierarchical working memory and context compression.

The function below shows an example of meta-agent's improvement on CCA's prompts. The part emphasized here is a {failure message} that gets raised when the best match is not exact. This message was iteratively refined by the meta-agent to be maximally actionable for an LLM: it shows the closest match, provides a minimal unified diff, and gives explicit constraints that force the agent to keep using \texttt{<file\_edit>} and to repair the \texttt{<find>/<find\_after>} tag into an exact \texttt{<line\_number>|<exact\_line\_content>} format rather than switching tools or attempting unsafe workarounds.

\begin{lstlisting}[language=Python,caption={Excerpt of the file-edit chunk matcher; non-prompt logic is collapsed as "\texttt{...}", while the meta-agent-refined prompt is shown verbatim.},label={lst:file_edit_prompt_excerpt}]
def _get_matched_chunk(...):
    # ... 
    
    if similarity == 1.0: 
        return matched_chunk

    # ... 

    # --- meta-agent-refined prompt begins
    raise ValueError(
        dedent(
            """\
        No exact occurrence found for the search string you provided.
        
        # ... shows match and diff
        
        ACTION REQUIRED: Please update your `<find>` or `<find_after>` tag to 
        match the exact content in the file with `<line_number>|
        <exact_line_content>` format.
        
        IMPORTANT: YOU MUST CONTINUE USING <file_edit> tag until successful. 
        DO NOT attempt alternative approaches such as:
        - Creating a new file to override the existing one
        - Using command line tools (e.g., `sed`, `awk`, etc.)
        
        Continue refining your `<find>` or `<find_after>` tag until it exactly 
        matches the file content.
        """
        ).format(...)
    )
    # --- meta-agent-refined prompt ends
\end{lstlisting}

\subsection{Evaluations on Long-context Reasoning}

\subsubsection{Context Management}

To quantify the impact of hierarchical working memory and context compression, we evaluate CCA on a subset of SWE-Bench-Pro, where both variants (with and without context management) successfully produced executable solutions.\footnote{Without any context control, many trajectories exceed model token limits and fail to complete, hence the restricted subset.} 
Results in \Cref{tab:context_mgmt} demonstrate a clear improvement in problem resolution when hierarchical memory and context compression are enabled. For Claude 4 Sonnet, advanced context management improves Resolve@1 from $42.0$ to $48.6$ on this subset (a +6.6 performance gain). On Claude 4.5, the improvement between the no-context-management and advanced variants is smaller, but both substantially outperform the simple tool-use configuration.  
This supports the hypothesis that structured context compression not only prevents overflow but also improves reasoning quality by enforcing periodic consolidation of long-horizon plans.
To ensure a controlled comparison, we use the same backbone LLM for both context compression and the main orchestrator agent across all experiments; see detailed study in the Appendix~\ref{app:context}.

\subsubsection{Endless-Read Robustness}

We further analyze CCA's robustness under tasks that require editing multiple files. Each SWE-Bench-Pro task is grouped by the number of modified files (``edited-file bucket''), and we measure the Resolve Rate within each group. As shown in \Cref{tab:endless_read}, the agent maintains stable performance across varying edit volumes, with only moderate regression when more files are touched. The degradations likely stem from cumulative localization uncertainty and compounding diffs, suggesting future work on finer-grained diff validation and multi-file dependency tracking.
Overall, these results show that CCA’s hierarchical memory and context compression yield substantial gains in both efficiency and robustness for long-context reasoning.

\begin{table}[h]
\centering
\begin{tabular}{l c c}
\toprule
\textbf{Edited Files} & \textbf{Resolve Rate (Pass@1)} & \textbf{\# Samples} \\
\midrule
1--2 files  & 57.8 & 294 \\
3--4 files  & 49.2 & 203 \\
5--6 files  & 44.1 & 86  \\
7--10 files & 52.6 & 38  \\
10+ files   & 44.4 & 18  \\
\bottomrule
\end{tabular}
\caption{CCA's resolve rate on SWE-Bench-Pro as a function of the number of files modified. 
Performance remains robust even for multi-file refactoring scenarios.}
\label{tab:endless_read}
\end{table}

\subsection{Evaluations on Long-term Memory}

We next study CCA’s \emph{note-taking} module, designed to accumulate durable cross-session memory. Unlike transient hierarchical working memory, the note-taking agent asynchronously summarizes each session into structured Markdown notes with multiple steps of reasoning, which capture both successful strategies and failure cases. This persistent ``memory'' is then available for retrieval in subsequent tasks, supporting test-time self-improvement.

Since no public benchmark explicitly evaluates memory in coding agents, we assess CCA’s memory module by running it on two consecutive passes, i.e., with memory maintained, of SWE-Bench-Pro instances. During the first run, the note-taking agent analyzes each trajectory and produces persistent notes for 151 instances—skipping cases where no meaningful insight can be distilled. We then rerun exactly these 151 tasks, providing CCA with the corresponding note directory to measure how prior experience improves efficiency and solution quality.
For Run 1: We execute the task from scratch (no context editing either); use note taker agent to write  down notes. For Run 2: We pass the notes from Run~1 to CCA and rerun.

\begin{table}[h]
\centering
\small
\begin{tabular}{l l l l}
\toprule
\textbf{Trial} & \textbf{Avg.\ Turns ($\downarrow$)} & \textbf{Avg.\ Token Cost ($\downarrow$)} & \textbf{Resolve Rate (Pass@1, $\uparrow$)} \\
\midrule
Run 1 & 64 & 104k & 53.0 \\
Run 2 & 61 \textcolor{red}{\textbf{(-3)}} & 93k \textcolor{red}{\textbf{(-11k)}} & 54.4 \textcolor{red}{\textbf{(+1.4)}} \\
\bottomrule
\end{tabular}
\caption{CCA performance across repeated runs using notes. For Run 1, all tasks are processed from scratch, and notes are taken and stored. For Run 2, each task resolving session is accompanied by the taken notes as the long-term memory. Token cost excludes system prompt tokens; the underlying model is Claude 4.5 Sonnet.}
\label{tab:note_ablation}
\end{table}

Cumulative note-taking reduces the iteration turns (from 64 to 61) and the token cost (from $104$k to $93$k), and also yield improvements on resolve rate (from 53\% to 54.4\%). These gains indicate that the notes distilled in the first run capture actionable, reusable knowledge. In effect, the note-taking system provides CCA with a lightweight form of \emph{cross-session learning}, enabling more efficient reasoning and more reliable patch generation in subsequent attempts. A detailed example of the notes produced by the note-taking agent is provided in Appendix~\ref{app:notes}.

\subsection{Comparison with Open-Sourced Scaffolds on SWE-Bench-Verified}

We further conduct evaluations on the SWE-Bench-Verified benchmark \citep{swebench} to compare CCA against existing open-source scaffolds.\footnote{As of Dec 2025, OpenHands remains the strongest open-sourced coding agent on SWE-Bench-Verified, reported from SWE-Bench's official leaderboard.} Using Claude 4 Sonnet, CCA achieves a Resolve Rate of \textbf{74.6\%}, exceeding the strongest open-source system (OpenHands) under identical backbone conditions and outperforming a mini-SWE-Agent variant that relies on the more capable Claude 4.5 Sonnet model. These results reinforce the central role of agentic scaffolding: improved orchestration, memory handling, and tool-use abstractions can close—or even surpass—the gap introduced by differences in backbone model capability. We also observe that SWE-Bench-Verified is sensitive to Claude’s internal thinking budget; a detailed analysis appears in Appendix~\ref{app:thinking}.

\begin{table}[h]
\centering
\begin{tabular}{l l c}
\toprule
\textbf{Backbone Model} & \textbf{Scaffold} & \textbf{Resolve Rate (Pass@1)} \\
\midrule

\multirow{3}{*}{Claude 4 Sonnet}
    & SWE-Agent            & 66.6 \\
    & OpenHands \citep{openhands}            & 72.8 \\
    & \textbf{CCA}         & \textbf{74.6} \\
\midrule

Claude 4.5 Sonnet
    & mini-SWE-Agent       & 70.6 \\
\bottomrule
\end{tabular}
\caption{CCA performance on SWE-Bench-Verified. CCA matches the best open-source framework (OpenHands) under the same Claude 4 Sonnet backbone, and outperforms a mini-SWE-Agent variant even when that variant uses a stronger Claude~4.5~Sonnet backbone.}
\label{tab:swe_verified}
\end{table}

\section{Related Work on Coding Agents}

A range of agent designs have been proposed to improve LLM-based software engineering. SWE-Agent~\citep{sweagent} is first proposed as a foundational system showing that an LLM paired with a small tool set (file editing, command execution, testing, etc.) can iteratively interact with real repositories to resolve issues. Subsequent work refines this paradigm in different directions. Live-SWE-Agent~\citep{xia2025live} studies test-time self-evolution, adapting strategies and occasionally updating prompts, tools, or configuration mid-run based on partial progress. Satori-SWE~\citep{satoriswe} explores population-based evolution at inference time, evolving multiple agent instances or candidates to improve sample efficiency via systematic refinement. In contrast, Agentless~\citep{xia2024agentless} argues for reducing agent complexity by replacing the open-ended loop with a fixed three-stage pipeline (localization, patch generation, test-case generation), achieving strong results on SWE-Bench Lite. Beyond academic prototypes, open-source platforms also shape the ecosystem: OpenHands~\citep{openhands} provides a community toolkit with a unified API for file I/O and code execution and implements a ReAct-style planner over popular base models. More related work can be found at Appendix \ref{sec.more_rw}.

\section{Conclusion}

We presented the \textbf{Confucius Code Agent (CCA)}, a coding agent for large-scale codebases. Built on the \textbf{Confucius SDK}, CCA separates and optimizes \emph{Agent Experience (AX)}, \emph{User Experience (UX)}, and \emph{Developer Experience (DX)}, enabling multi-step reasoning with modular tools, structured memory, and interpretable traces. Across public benchmarks and real-world settings, CCA shows that \emph{agentic scaffolding}—orchestration, memory, and tool abstractions—can matter as much as, or more than, the backbone model. The SDK’s hierarchical memory, context compression, and persistent notes support long-horizon stability, while extensions and the meta-agent facilitate rapid adaptation to new tools and workflows.

More broadly, Confucius SDK establishes a new foundation for AI agent research. Its modular architecture invites experimentation: from studying long-context reasoning and long-term memory, to exploring test-time adaptation, to integrating reinforcement learning with structured trajectory traces. We hope this framework accelerates progress toward AI developers that are powerful, interpretable, and continuously improving, bridging the persistent gap between research prototypes and the demands of real-world software engineering.

\clearpage
\newpage

\bibliographystyle{assets/plainnat}
\bibliography{ref}

\clearpage
\newpage

\beginappendix

\section{Additional Related Work}
\label{sec.more_rw}

\subsection{Benchmarking Coding Agents} 
The past two years have seen the emergence of comprehensive benchmarks to evaluate autonomous code-writing and code-fixing agents on realistic tasks. One prominent example is SWE-Bench \citep{swebench}, which poses real-world GitHub issues and provides the full repository context; an agent succeeds by producing a patch that passes all project tests. 
It has since evolved into a family of benchmarks targeting different scenarios: for instance, variants like SWE-Bench-Multilingual \citep{yang2025swesmith} and SWE-Bench-Multimodal \citep{yang2025swebench} extend the evaluation to codebases with multiple programming languages and to tasks that involve not only code but other modalities (such as modifying documentation or configurations), respectively. An expanded challenge, SWE-Bench Pro \citep{sbp}, was released to assess long-horizon problem solving: it includes complex, enterprise-level issues that may require dozens of files to be modified across a codebase. These benchmarks have become a driving force for the community, with public leaderboards spurring rapid progress. Beyond bug-fixing, entirely new benchmarks are probing other dimensions of software work. SWE-fficiency \citep{ma2025swe} is a recent benchmark that challenges agents to optimize the runtime performance of real codebases given defined workloads. 
% In summary, a rich benchmarking ecosystem now exists for software engineering AI, and it highlights a gap between what current LLM agents can do and the full spectrum of a professional developer’s skill set.

\subsection{Large-scale Software Engineering}
Modern software engineering at scale has driven interest in AI assistance that can handle massive codebases and performance-critical systems. Potvin and Levenberg’s seminal description of Google’s single vast code repository illustrates the challenges and benefits of the monorepo model \citep{potvin2016google}. This approach centralizes billions of lines of code, enabling unified tooling and refactoring, but it demands automated support for code discovery, understanding, and consistent changes at scale. Recent LLM-based systems are beginning to tackle such issues. For instance, Lin et al. introduce ECO, an LLM-driven code optimizer designed for warehouse-scale computers \citep{lin2025eco}. ECO leverages a code-generating model to suggest performance improvements in large distributed software, aiming to reduce runtime and resource usage while preserving correctness. Early results show that AI-powered optimization can uncover non-trivial efficiency gains in complex systems, hinting at a future where coding agents assist not only in writing code but also in optimizing and maintaining it across ultra-large codebases. The combination of monorepo development and LLM-based tools like ECO underscores a trend toward holistic scale: treating an entire organization’s code as a single evolvable system, with AI agents providing the intelligence to manage global changes, dependency analysis, and performance tuning in ways humans alone could not easily scale. This context also motivates advanced context management techniques – instead of feeding billions of lines directly into an LLM, agents must learn to retrieve and focus on the relevant project fragments, a theme that connects to memory and tool-use innovations discussed later.

\subsection{Training LLMs for Software Engineering} 
SWE-Gym \citep{pan2024training} provides the first publicly available executable environment tailored for real-world software engineering tasks: it bundles complete Python repositories with dependencies, unit tests, and realistic issue descriptions, enabling agents to propose patches which can be validated via execution. Later, SWE-Smith \citep{yang2025swesmith} generalizes the idea: given any Python repository, it automatically generates hundreds to thousands of new bug-fix or issue-resolution tasks by perturbing code or simulating realistic faults, producing a dataset of around 50,000 instances across 128 GitHub projects. Training on this large-scale synthetic data significantly improves agent performance on benchmark tasks, indicating that domain-specific, execution-aware fine-tuning is critical for bringing coding agents closer to real-world software engineering demands.
Recent research have explored reinforcement learning for code agents. SWE-RL \citep{wei2025swe} takes advantage of the abundant software evolution data in open-source repositories – commit histories, diff patches, and issue resolutions – and uses these as implicit demonstrations to fine-tune an LLM via RL. The insight is that by learning from how human developers iteratively improve code over many commits, an agent can internalize more realistic problem-solving trajectories. Early results from SWE-RL show improved reasoning and success rates on long-horizon software tasks, as the RL-trained model learns to recover intermediate reasoning steps (e.g. the decision to run tests or check certain files) that are often needed for complex bug fixes.

\section{Context Summarization Ablations}
\label{app:context}
The architect agent summarizes its working memory during execution when it exceeds given threshold; we evaluate whether the quality of context summarization affects downstream agent performance.
In this ablation, we fix the summary \emph{format}, and approximately keep the summary \emph{length}, and \emph{summarization cadence} in comparable ranges, and vary only the LLM used to generate summaries.
All runs use the same main orchestrator (Claude~4~Sonnet) and are evaluated on SWE-Bench Pro.
We compare Claude~3.5~haiku and Claude~4~Sonnet as summarizers.
haiku is chosen as the ``weak'' model because it reliably produces well-formed, schema-compliant summaries, while being substantially weaker in reasoning ability: on SWE-Bench, haiku achieves 40.6\% versus Sonnet’s 72.7\%, and lacks extended multi-step reasoning, hence this choice avoids confounding effects from malformed summaries or instruction non-compliance.

We select 50 long-context instances in which summarization is triggered at least once during execution (up to five times in the longest cases).
Under identical conditions, haiku-based summarization resolves 22/50 instances, while Sonnet-based summarization resolves 26/50. The observed gap directly demonstrates that higher-fidelity, reasoning-capable summaries provide better guidance for subsequent planning and lead to higher final resolve rates.

To illustrate why the quality of the summarization matters, here we provide summary examples of \texttt{instance\_gravitational\_\_teleport-db89206db6c2969266e664c7c0fb51b70e958b64}.

Here is the context summary of Claude Haiku 3.5: 
\begin{tcolorbox}[colback=white,  % background color
                   colframe=black,    % frame color
                   coltext=metablue,      % text color
                   breakable,
                   fontupper=\scriptsize\fontfamily{pcr}\selectfont]  % font style and size
\begin{verbatim}
<summary>
 [CONVERSATION CONTEXT]
- Initial Requirements: Replace fatal error handling in TSH CLI commands with 
error return mechanisms
- Scope Changes: Modify command handlers to return errors instead of calling 
utils.FatalError
- User Preferences: Maintain existing logging and error handling patterns while 
improving test environment flexibility

[TECHNICAL DECISIONS]
- Architecture Decisions: Convert command handler functions to return errors
- Technology Stack: Go language, Teleport client library
- Design Patterns: Error wrapping using trace.Wrap(), trace.BadParameter()
- APIs/Interfaces: Modify function signatures in tool/tsh/tsh.go and 
lib/client/api.go

[IMPLEMENTATION PROGRESS]
- Completed Work: 
  * Added SSOLoginFunc type
  * Modified ssoLogin method
  * Updated onSSH, onPlay command handlers
  * Started modifying onLogin and onLogout
- Current State: Partial implementation of error handling refactoring
- failed Attempts: Initial attempts to replace utils.FatalError with simple 
return statements
- Debugging History: systematic approach to replacing error handling across 
multiple files

[TECHNICAL Details]
- Data Structures: CLIConf, Config structs modified
- Algorithms: Error propagation and wrapping
- Edge Cases: Handling various error scenarios in CLI commands
- Performance Considerations: Minimal performance impact, improved testability

[OUTSTANDING ITEMS]
- Known Issues: Incomplete error handling in complex functions like onLogin
- Open Questions: How to handle early returns in long functions
- TODOs: Complete error handling for remaining command handlers
</summary>
<plan>
 <thinking>
  - Problem Analysis: Complex error handling refactoring across multiple files
- Impact Assessment: Improved testability and error management in Teleport CLI
- Solution Strategy: systematic conversion of command handlers to return errors
 </thinking>
 <issues>
  1. utils.FatalError calls prevent programmatic error handling
2. Complex functions with multiple error points need careful refactoring
3. Function signatures must change to support error returns
 </issues>
 <next_step sequence_num="1">
  Complete error handling conversion for onLogin function
- Replace all utils.FatalError calls with return trace.Wrap(err)
- Handle early return cases by returning nil or specific errors
- Ensure function signature returns error
 </next_step>
\end{verbatim}
\end{tcolorbox}
Here is the summary of Claude Sonnet 4:
\begin{tcolorbox}[colback=white,  % background color
                   colframe=black,    % frame color
                   coltext=metablue,      % text color
                   breakable,
                   fontupper=\scriptsize\fontfamily{pcr}\selectfont]  % font style and size
\begin{verbatim}
<summary>
 [CONVERSATION CONTEXT]
- Initial Requirements: Fix TSH login and proxy address handling for test 
environments by making CLI commands return errors instead of exiting, supporting
mock SSO login injection, and using runtime-assigned addresses for services 
bound to `:0`
- Scope Changes: None - requirements remained consistent throughout
- User Preferences: systematic approach with todo.md planning, focus on minimal 
changes, backward compatibility

[TECHNICAL DECISIONS]
- Architecture Decisions: Chose to implement error returns instead of process 
termination for all command handlers, added pluggable SSO login interface
- Technology Stack: Go, existing Teleport codebase structure maintained
- Design Patterns: Used function type for SSO login injection, trace.Wrap for 
error propagation
- APIs/Interfaces: Defined SSOLoginFunc type with specific signature matching 
existing ssoLogin method

[IMPLEMENTATION PROGRESS]
- Completed Work: 
  * Phase 1: Added SSOLoginFunc type, MockSSOLogin field to Config struct, 
mockSSOLogin field to CLIConf struct
  * Partial Phase 2: Updated refuseArgs helper and several command handlers 
(onLogout, onListNodes, onListClusters, onSSH) to return errors
- Current State: In middle of Phase 2, about to update Run function in Phase 3
- failed Attempts: None significant
- Debugging History: None required yet

[TECHNICAL Details]
- Data Structures: Added SSOLoginFunc type, MockSSOLogin field to Config, 
mockSSOLogin field to CLIConf
- Algorithms: Error propagation using trace.Wrap, replacing utils.FatalError and
os.Exit calls
- Edge Cases: Handling exit status preservation in SSH command, ambiguous host 
error handling
- Performance Considerations: Minimal impact, just changing error handling flow

[OUTSTANDING ITEMS]
- Known Issues: None identified yet
- Open Questions: Need to complete remaining command handlers and service 
address binding logic
- TODOs: Complete Phase 2 (remaining command handlers), Phase 3 (Run function), 
Phase 4 (SSO mock), Phase 5 (service binding), Phase 6 (validation)
</summary>
<plan>
 <thinking>
  - Current State: Successfully implemented Phase 1 (new types/interfaces) and 
partially completed Phase 2 (error handling for some command handlers)
- Architecture Implications: The error handling changes are fundamental and 
affect the entire CLI flow
- Risk Assessment: Need to ensure all command handlers are updated consistently 
and the Run function properly handles all error cases
 </thinking>
 <next_step sequence_num="1">
  Update the Run function signature to return error and modify the switch 
statement to capture errors from all command handlers, including updating the 
refuseArgs call for logout and replacing utils.FatalError with proper error 
return
 </next_step>
\end{verbatim}
\end{tcolorbox}

The summary format and token count are almost identical between the 2 summaries, while Sonnet 4 appears much better in quality. The "[CONVERSATION CONTEXT]" of sonnet preserves maximum information: ``Fix TSH login + proxy handling for test environments by returning errors instead of exiting...", while the haiku summary ``Replace fatal error handling with error returns." loses important keywords.
Most notably, Sonnet shows phases of work: \textbf{it specifies state of the system (Phase 1 done, Phase 2 mid-way, next is Run function)}, making progress legible and actionable..

\section{Meta-Agent}
\subsection{Implementation Details}
\label{sec.meta-agent-details}
\begin{figure*}[t]
    \centering
    \includegraphics[width=\textwidth]{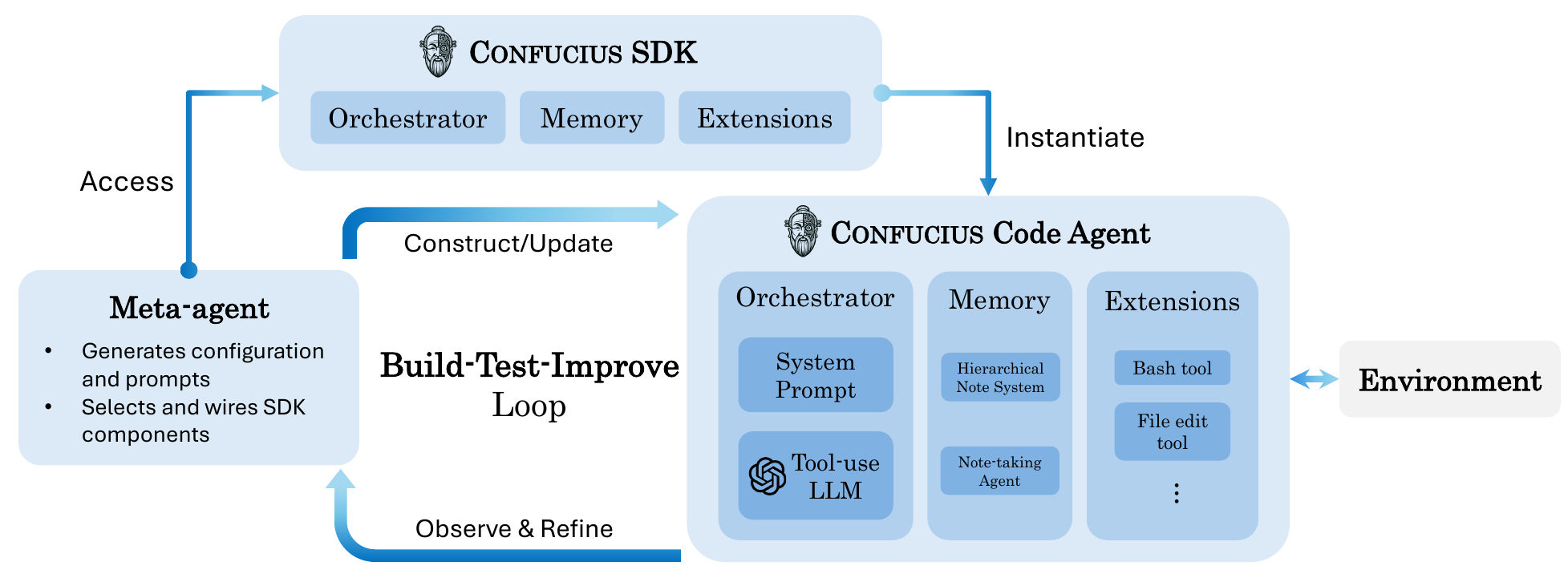}
    \caption{\textbf{Meta-agent build–test–improve loop.} The Confucius Code Agent (CCA) is a specific agent built on top of the Confucius SDK, with the help of the meta-agent. The Meta-agent synthesizes agent configurations, wires together orchestrator components and extensions, evaluates candidate agents on representative tasks, and iteratively refines prompts and tool-use policies based on observed failures.}
    \label{fig:meta-agent}
\end{figure*}

As \Cref{fig:meta-agent} shows, the Meta-agent is implemented as an extra agent that interactively constructs new agents from high-level specifications based on the Confucius SDK. A developer begins by describing, in natural language, what the target agent should do and under what constraints (e.g., ``an agent that triages CI failures for our monorepo'' or ``a refactoring agent with read-only access to production configs''). The Meta-agent then generates a structured configuration form that asks for more concrete requirements: repository scope, latency or safety constraints, which existing extensions (file editing, bash, code search, etc.) to attach, and what evaluation tasks or test suites should be used. After the user confirms this specification, the Meta-agent automatically (i) synthesizes the agent’s configuration and prompts and (ii) wires in the selected extensions and memory policies. We present a brief illustration of this user interface in Appendix~\ref{sec.meta-agent-dev-cycle}.

Importantly, the Meta-agent also automates \emph{testing and debugging} of the newly created agent. Using the same SDK runtime, it spins up the candidate agent locally, drives it on a suite of regression tasks (e.g., representative GitHub issues or internal tickets), and observes the agent’s outputs, logs, and tool traces. When failures or undesirable behaviors are detected, such as brittle tool selection, incorrect file-edit patterns, or poor recovery from compiler errors, the Meta Agent proposes concrete modifications to prompts, extension configurations, or even new tool wrappers. These patches are applied to the agent, and the test loop is rerun, yielding a ``build-test-improve'' process that incrementally improves the agent until target metrics are met. The same mechanism can be invoked not only to build new agents, but also to assist in designing and debugging new tools that plug into the extension layer.

\subsection{Development Cycle}
\label{sec.meta-agent-dev-cycle}

\begin{figure}[h]
  \centering
  \includegraphics[width=\textwidth]{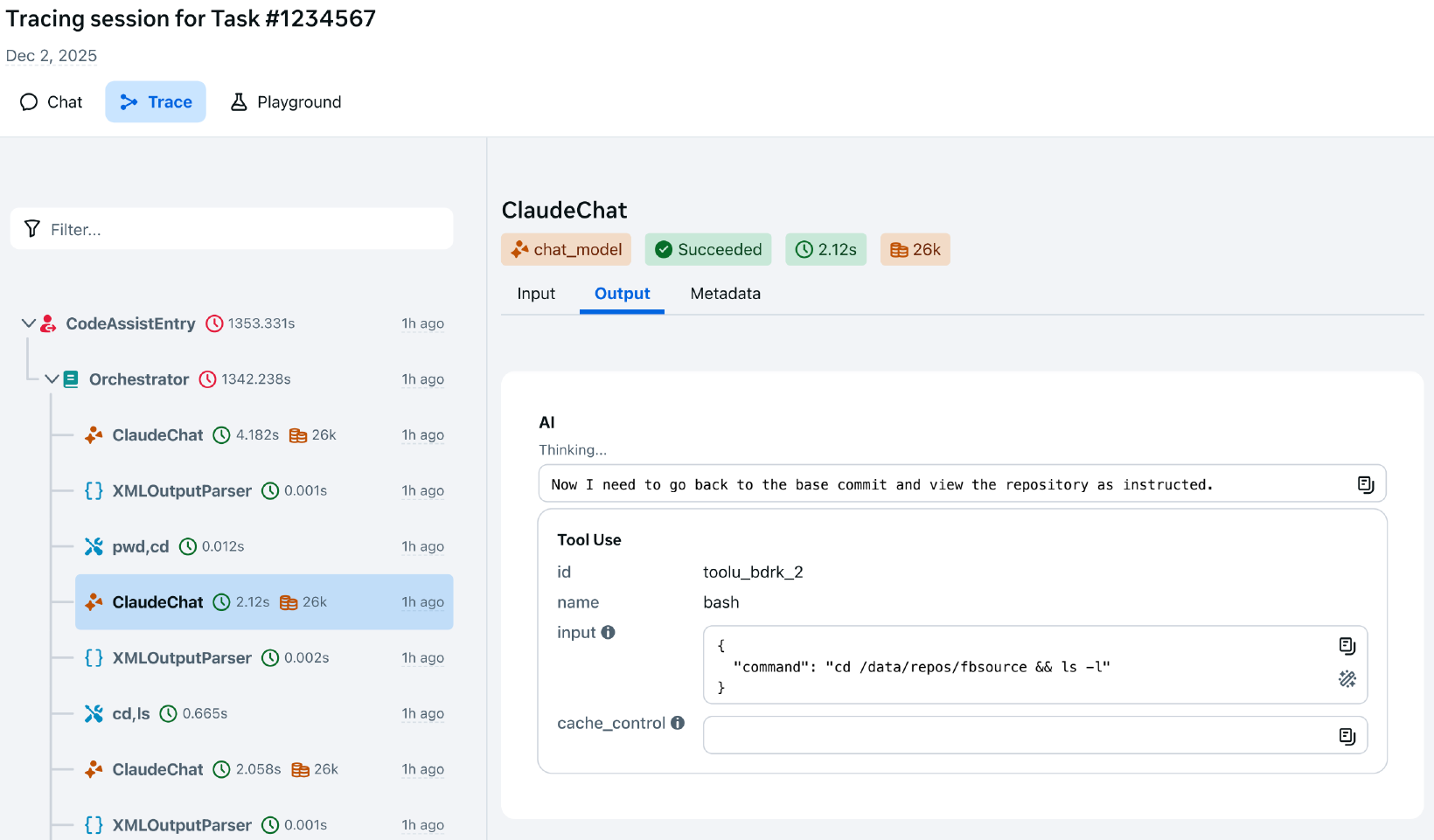}
  \caption{CCA Trace UI with call stack visualization and tool invocation Details, providing developers with detailed visibility into agent execution, showing the hierarchical call stack, latency metrics, token usage, and tool invocations for debugging and performance optimization. }
  \label{fig:trace_ui_v2}
\end{figure}

The Confucius SDK promotes an easy-to-use \textbf{agent development cycle} where the meta-agent assist in onboarding and refining other agents.  
This iterative process of build-test-improve loop is further supported by a full suite of developer tools:  
\begin{itemize}[leftmargin=*]
\item \textbf{Onboarding Experience:} Meta-agent provides an easy-to-use interface for developers to create an agent from scratch, the onboarding experience offers multiple coding agent templates (orchestrators + extensions) and a multi-turn Q\&A session with developer to clarify requirements, it also offers E2E testing capabilities, where user can on-the-fly build test cases and optimize the agent code/prompt.
\item \textbf{Trace UI:} fine-grained visualization of call stacks, tool interactions, and memory flows, as shown in Figure~\ref{fig:trace_ui_v2}
\item \textbf{Playground:} an interactive environment for prompt refinement and parameter tuning;
\item \textbf{Eval UI:} built-in support for regression tests, A/B comparisons, and benchmark evaluations;
\item \textbf{Centralized agent management:} a unified interface for developing, integrating, deploying, and monitoring agents at scale.
\end{itemize}

\section{Thinking Budget Scaling}
\label{app:thinking}
We define the \emph{thinking budget} as the maximum number of reasoning or chain-of-thought ``thinking'' tokens that the LLM is permitted to generate before producing a response. Recent work shows that reasoning length can be controlled by prompting or internal budget-aware mechanisms. At inference time, the model can be guided either by: (i) prompt instructions like ``use at most $X$ tokens of reasoning''~\citep{han2025token}; (ii) control tokens inserted periodically that signal remaining budget to the model during generation~\citep{wen2025budgetthinker}; or (iii) a predictor estimating task complexity and adjusting budget dynamically~\cite{li2025steering}. 

Anthropic's Claude model series expose a \texttt{thinkingBudget} parameter at inference time to directly cap the reasoning tokens. We run our CCA agent using Claude 4 Sonnet with identical setup except varying its thinking budget in three settings: $8$k, $16$k, and $32$k tokens. We record Resolve Rate on a subset of SWE-Bench-Verified.

\begin{table}[h]
\centering
\begin{tabular}{l c}
\toprule
\textbf{Thinking Budget} & \textbf{Resolve Rate (Pass@1)} \\
\midrule
8k   & 67.3 \\
16k  & 68.4 \\
32k  & 68.7 \\
\bottomrule
\end{tabular}
\caption{CCA Resolve Rate as a function of thinking budget on SWE-Bench-Verified subset.}
\label{tab:thinking_budget}
\end{table}

We observe diminishing returns beyond $16$k thinking tokens. Note that the \texttt{thinkingBudget} parameter cannot precisely control the internal thinking trace length of the Claude model, and that during inference, Claude only returns a summarized version of the reasoning without exposing full traces. As a result, we do not have the exact scaling curve of Resolve Rate with respect to the true thinking length on SWE-Bench-Verified.

\section{Example Execution Trace}
\label{app:trace}

Below, we show an example execution trace when running CCA on a SWE-Bench-Pro example: \texttt{instance\_tutao\_\_tutanota-da4edb7375c10f...}

\begin{tcolorbox}[colback=white,  % background color
                   colframe=black,    % frame color
                   coltext=metablue,      % text color
                   breakable,
                   fontupper=\scriptsize\fontfamily{pcr}\selectfont]  % font style and size
\begin{verbatim}
Running Confucius Code Agent with prompt from file: <PROMPT_FILE>

<SYSTEM>
Viewing directory at `<WORKDIR>`
Directory content viewed at `<WORKDIR>`, depth: 2, show_hidden: False

<AI>
The repository is in `<REPO_ROOT>`. Let me explore its structure:

<SYSTEM>
Viewing directory at `<REPO_ROOT_NAME>`
Directory content viewed at `<REPO_ROOT_NAME>`, depth: 2, show_hidden: False

<AI>
Let me search for `EntityRestClient` and related files:

<SYSTEM>
Validating command `find <REPO_ROOT> -type f -name "*.ts" | xargs grep -l "EntityRestClient" ... | head -20`
Running command in `<WORKDIR>`:
  find <REPO_ROOT> -type f -name "*.ts" | xargs grep -l "EntityRestClient" ... | head -20
Output:
  <REPO_ROOT>/src/api/common/EntityClient.ts
  <REPO_ROOT>/src/api/worker/rest/EntityRestClient.ts
  <REPO_ROOT>/src/api/worker/facades/BlobAccessTokenFacade.ts
  ...
Viewing file at `EntityRestClient.ts`
File content viewed at `EntityRestClient.ts`, lines: None - None

<AI>
Now let me search for `requestReadTokenArchive` and `requestReadTokenBlobs` in the codebase:

<SYSTEM>
Validating command `grep -rn "requestReadTokenArchive\|requestReadTokenBlobs" <REPO_ROOT>/src --include="*.ts"
...`
Running command in `<WORKDIR>`:
  grep -rn "requestReadTokenArchive\|requestReadTokenBlobs" <REPO_ROOT>/src --include="*.ts" ...
Output:
  <REPO_ROOT>/src/api/worker/facades/BlobAccessTokenFacade.ts:... requestReadTokenBlobs(...)
  <REPO_ROOT>/src/api/worker/facades/BlobAccessTokenFacade.ts:... requestReadTokenArchive(...)
  <REPO_ROOT>/src/api/worker/rest/EntityRestClient.ts:... requestReadTokenArchive(..., listId)
  ...

<SYSTEM>
Viewing file at `BlobAccessTokenFacade.ts`
File content viewed at `BlobAccessTokenFacade.ts`, lines: None - None

<AI>
Now let me look at the `createBlobAccessTokenPostIn` and related types to understand the structure:

<SYSTEM>
Validating command `grep -rn "createBlobAccessTokenPostIn\|BlobAccessTokenPostIn" <REPO_ROOT>/src 
--include="*.ts" ... | head -20`
Running command in `<WORKDIR>`:
  grep -rn "createBlobAccessTokenPostIn\|BlobAccessTokenPostIn" <REPO_ROOT>/src --include="*.ts" ... | head -20
Output:
  <REPO_ROOT>/src/api/entities/storage/TypeRefs.ts:... export type BlobAccessTokenPostIn = { ... }
  ...
Viewing file at `TypeRefs.ts`
File content viewed at `TypeRefs.ts`, lines: None - None

<AI>
(Problem analysis)
- `requestReadTokenArchive` / `requestReadTokenBlobs` currently require `archiveDataType`
- `EntityRestClient` hardcodes `ArchiveDataType.MailDetails`
- Requirement: allow `archiveDataType = null` for owned archives, keep behavior unchanged for non-owned
archives
...

<AI>
Now let me check the git status to see if there are any changes already made:

<SYSTEM>
Validating command `cd <REPO_ROOT> && git status`
Command rejected (disallowed): `cd ...`
Validating command `git -C <REPO_ROOT> status`
Running command in `<WORKDIR>`:
  git -C <REPO_ROOT> status
Output:
  HEAD detached from <COMMIT_SHA>
  nothing to commit, working tree clean

<AI>
This is a moderately complex task that involves changes to multiple files. Let me plan this out systematically.

<SYSTEM>
Writing to memory node 'todo.md'...
Created memory node 'todo.md' with ... characters and tags: ...

<AI>
Now let me create a reproduction script to understand the current behavior:

<SYSTEM>
Creating file at `reproduce.py`
File created successfully at `reproduce.py`

<AI>
Now let's run the reproduction script to confirm the issue:

<SYSTEM>
Validating command `python3 <REPO_ROOT>/reproduce.py`
Running command in `<WORKDIR>`:
  python3 <REPO_ROOT>/reproduce.py
Output:
  Test 1: ... archiveDataType allows null?  FAIL
  Test 2: ... method signatures accept null? FAIL
  Test 3: ... EntityRestClient passes null?  FAIL
  Some tests failed. The issue needs to be fixed.

<AI>
Implementing the fix.

<SYSTEM>
Replacing content in file at `TypeRefs.ts`
Diff:
  - archiveDataType: NumberString;
  + archiveDataType: NumberString | null;

<SYSTEM>
Replacing content in file at `BlobAccessTokenFacade.ts`
Diff:
  - requestReadTokenBlobs(archiveDataType: ArchiveDataType, ...)
  + requestReadTokenBlobs(archiveDataType: ArchiveDataType | null, ...)
  - requestReadTokenArchive(archiveDataType: ArchiveDataType, ...)
  + requestReadTokenArchive(archiveDataType: ArchiveDataType | null, ...)

<SYSTEM>
Replacing content in file at `EntityRestClient.ts`
Diff:
  - requestReadTokenArchive(ArchiveDataType.MailDetails, listId)
  + requestReadTokenArchive(null, listId)

<SYSTEM>
Replacing content in file at `EntityRestClient.ts`
Diff:
  - import { ArchiveDataType } from ".../TutanotaConstants.js"
  + (removed unused import)

<AI>
Now let's run the reproduction script again to verify the fix:

<SYSTEM>
Validating command `python3 <REPO_ROOT>/reproduce.py`
Running command in `<WORKDIR>`:
  python3 <REPO_ROOT>/reproduce.py
Output:
  Test 1: ... allows null, PASS
  Test 2: ... signatures accept null, PASS
  Test 3: ... passes null, PASS
  All tests passed! The fix is working correctly.

<AI>
Verifying diffs and committing.

<SYSTEM>
Validating command `git -C <REPO_ROOT> diff --name-only`
Running command in `<WORKDIR>`:
  git -C <REPO_ROOT> diff --name-only
Output:
  src/api/entities/storage/TypeRefs.ts
  src/api/worker/facades/BlobAccessTokenFacade.ts
  src/api/worker/rest/EntityRestClient.ts

<SYSTEM>
Validating command `git -C <REPO_ROOT> add -A && git -C <REPO_ROOT> commit -m "<COMMIT_MSG>"`
Running command in `<WORKDIR>`:
  git -C <REPO_ROOT> add -A && git -C <REPO_ROOT> commit -m "<COMMIT_MSG>"
Output:
  Committer: <ANON_NAME> <<ANON_EMAIL>>
  3 files changed, ... insertions(+), ... deletions(-)

<AI>
Done. Changes preserve existing behavior for non-owned archives, while allowing owned-archive read token 
requests without archiveDataType.
\end{verbatim}
\end{tcolorbox}

\section{Example Notes for Long-term Memory}
\label{app:notes}
Here is an example structured note hierarchy generated by note-taking agent for SWE-bench-Pro instance: \texttt{instance\_internetarchive\_\_openlibrary-f343c08f89c772f7ba...}.

\begin{tcolorbox}[colback=white,  % background color
                   colframe=black,    % frame color
                   coltext=metablue,      % text color
                   breakable,
                   fontupper=\scriptsize\fontfamily{pcr}\selectfont]  % font style and size
\begin{verbatim}
.
+-- projects
|   +-- openlibrary
|       +-- escaping_wildcards_in_infobase_queries.md
|       +-- multi_stage_author_matching_pipeline.md
|       +-- year_based_author_matching_strategy.md
+-- README.md
+-- shared
    +-- python
    |   +-- dict_copy_forgotten_field_update.md
    +-- string_manipulation
        +-- prefix_removal_empty_string_edge_case.md
\end{verbatim}
\end{tcolorbox}

The note is well organized into "shared" (ONLY for truly generic insights that apply across MANY projects) and "project" (For project-specific knowledge that primarily applies to ONE specific domain). The README.md summarized all the notes md files and provide instructions on how to navigate them. Here are some insights from \texttt{"escaping\_wildcards\_in\_infobase\_queries.md"}.

\begin{tcolorbox}[colback=white,  % background color
                   colframe=black,    % frame color
                   coltext=metablue,      % text color
                   breakable,
                   fontupper=\scriptsize\fontfamily{pcr}\selectfont]  % font style and size
\begin{verbatim}
---
id: escaping_wildcards_in_infobase_queries
title: Escaping Wildcards in Infobase Queries
description: How to handle asterisk characters in \\
Infobase queries to prevent unintended wildcard matching
keywords:
    - infobase
    - queries
    - wildcards
    - escaping
    - openlibrary
---

# Escaping Wildcards in Infobase Queries

## Problem Context

When searching for authors by name in OpenLibrary's Infobase, asterisk characters 
(`*`) in author names can cause unexpected behavior because they are treated as 
wildcards by the query system. For example, searching for an author named "John*" 
would match "John Smith", "Johnny", etc., rather than the literal name "John*". 


## The Solution

Escape asterisks in name fields using a backslash when performing exact or alternate 
name matches, but **intentionally preserve wildcards for surname matching queries**. 

### Code Example

```python
# In find_author() function:

# Escape asterisks for exact name matching
escaped_name = author["name"].replace("*", r"\*")

queries = [
    # Use escaped name for exact match
    {"type": "/type/author", "name~": escaped_name},
    # Use escaped name for alternate names
    {"type": "/type/author", "alternate_names~": escaped_name},
]

# For surname matching, intentionally use wildcards
if birth_year and death_year:
    surname = author['name'].split()[-1]
    queries.append({
        "type": "/type/author",
        "name~": f"* {surname}",  # Wildcard intentional here
        "birth_date~": f"*{birth_year}*",
        "death_date~": f"*{death_year}*",
    })
```

## Key Insights

1. **Context-Dependent Escaping**: Wildcards need to be escaped in some query contexts
(exact/alternate name matching) but preserved in others (surname matching).

2. **Backslash Escaping**: In Infobase queries, use `r"\*"` to escape asterisks. The
raw string `r` prefix prevents Python from interpreting the backslash.

3. **Query Order Matters**: The queries are tried in sequence. Exact matches (with
escaped wildcards) are tried first, then alternate names, then surname matching (with
intentional wildcards).

4. **Creating New Records**: When no match is found and a new author record must be
created, preserve the original name including any asterisks exactly as provided - 
don't use the escaped version.


## Related Files

- `/app/openlibrary/catalog/add_book/load_book.py` - `find_author()` function
- `/app/openlibrary/catalog/add_book/tests/test_load_book.py` - Tests for wildcard
behavior
\end{verbatim}
\end{tcolorbox}
These insights saves the agent a lot of effort to rediscover the issue in Run 2. 
Here are the insights from \texttt{prefix\_removal\_empty\_string\_edge\_case.md}.

\begin{tcolorbox}[colback=white,  % background color
                   colframe=black,    % frame color
                   coltext=metablue,      % text color
                   breakable,
                   fontupper=\scriptsize\fontfamily{pcr}\selectfont]  % font style and size
\begin{verbatim}
---
id: prefix_removal_empty_string_edge_case
title: Prefix Removal Empty String Edge Case
description: How to handle the edge case where removing a prefix leaves an empty string
keywords:
    - string
    - manipulation
    - edge
    - case
    - prefix
    - validation
---

# Prefix Removal Empty String Edge Case

## Problem

When removing a prefix from a string, you may end up with an empty string if the input
consists only of the prefix. This can cause unexpected behavior if downstream code
doesn't handle empty strings properly.

### Example Scenario

In OpenLibrary's honorific removal function, names like "Mr." or "Dr." consist 
entirely of an honorific:

```python
def remove_author_honorifics(author: dict[str, Any]) -> dict[str, Any]:
    raw_name: str = author["name"]
    
    if honorific := find_matching_honorific(raw_name):
        author["name"] = raw_name[len(honorific):].lstrip()
    return author
```

**Problem**: If `raw_name = "Mr."` and we remove "mr.", we get an empty string `""`,
which can cause issues in database queries or display logic.

## The Solution

Check if the result is empty after removing the prefix, and if so, return the original value unchanged:


```python
def remove_author_honorifics(author: dict[str, Any]) -> dict[str, Any]:
    """Remove honorifics from an author's name field."""
    raw_name: str = author["name"]
    
    # Check for exceptions first (e.g., "Dr. Seuss")
    if raw_name.casefold() in HONORIFC_NAME_EXECPTIONS:
        return author
    
    if honorific := next(
        (
            honorific
            for honorific in HONORIFICS
            if raw_name.casefold().startswith(honorific)
        ),
        None,
    ):
        new_name = raw_name[len(honorific):].lstrip()
        # If the name is only an honorific, return unchanged
        if not new_name:
            return author
        author["name"] = new_name
    return author
```

## Key Insights

1. **Check Before Assignment**: Store the result in a temporary variable and validate
it before assigning to the final location.

2. **Preserve Original on Invalid Result**: When validation fails (empty string),
return the original input unchanged rather than the invalid result.

3. **Common in Text Processing**: This pattern applies to any string manipulation
where you're removing parts of the string:
   - Removing file extensions
   - Removing URL protocols
   - Stripping whitespace/punctuation
   - Removing prefixes/suffixes
\end{verbatim}
\end{tcolorbox}

This documents a concrete edge case when solving the instance; this not only save the agent tokens in Run 2, but also avoid runs where in consecutive runs such edge cases could be missed hence fail the case.  

\section{Case Studies: Comparison with Claude Code}
\label{sec:case_studies}

In addition to standardized benchmarks, we conducted a controlled experiment using real GitHub issues from the PyTorch repository. These issues not only exemplify the complex challenges encountered in real-world production but also require deep domain specialist expertise. Hence, these issues reflect agent's robustness and generalization under specialist software engineering scenarios. Our experiment holds the model capabilities constant while varying only the agent framework. We compare \textbf{CCA} with \textbf{Claude Code (CC)}, a command-line tool developed by Anthropic that enables direct interaction with Claude models for coding tasks. Both frameworks utilize identical Claude Sonnet 4.5 models in environments with equivalent codebases and access to file manipulation, bash tools, and NVIDIA A100 80GB GPU resources. To compare the solution between CC and CCA, we have enlisted a few experts in this field to judge and compare the solutions created by the 2 agents.

\subsection{PyTorch-Bench}

To construct PyTorch-Bench, we scanned GitHub issues on the open-source PyTorch repository \citep{pytorch2025} from Jan 2025 to Jul 2025. We selected 8 issues that are reproducible on an NVIDIA A100 80 GB GPU and that provide actionable structure, including a detailed description, a reproduction script, and instructions for replication of the issue. Both agents receive the same system prompt, which instructs them to start from a clean commit, attempt to reproduce the issue first, stop if reproduction fails in the current environment, and verify any proposed fix. We show an example task below and discuss it in more detail in the following sections.

\begin{tcolorbox}[colback=white,  % background color
                   colframe=black,    % frame color
                   coltext=metablue,      % text color
                   breakable,
                   fontupper=\scriptsize\fontfamily{pcr}\selectfont]  % font style and size
\begin{verbatim}
Issue: RuntimeError: Expected curr_block->next == nullptr to be true,  when I call setSegmentStateToCheckpoint. 
(#161356)
URL: https://github.com/pytorch/pytorch/issues/161356

### Describe the bug

Hello, when I was using checkpoint state to implement shared output memory for two cudagraphs, an assert 
ERROR occurred:
curr_block->next == nullptr, in function setSegmentStateToCheckpoint.

```
# PyTorch version: 2.6.0+cu124

torch._C._cuda_setCheckpointPoolState(com_device, small_state, [], output1_new_storage)
RuntimeError: Expected curr_block->next == nullptr to be true, but got false.  
(Could this error message be improved?  If so, please report an enhancement request to PyTorch.)
'''

**This error only appears only when I set env `export PYTORCH_CUDA_ALLOC_CONF=expandable_segments:True`.
If I unset this env, the following case code executes successfully.**

This is my test code:
```
import gc
import torch
print(f"torch version is: {torch.__version__}.")
stream0 = torch.cuda.Stream()
torch.cuda.set_stream(stream0)

def tensor_metadata(x):
    return {
        "nbytes": x.untyped_storage().nbytes(),
        "data_ptr": x.untyped_storage().data_ptr(),
        "size": x.shape,
        "stride": x.stride(),
        "dtype": x.dtype,
        "device": x.device,
        "storage_offset": x.storage_offset(),}

def reconstruct_from_tensor_metadata(metadata):
    s = torch._C._construct_storage_from_data_pointer(
        metadata["data_ptr"], metadata["device"], metadata["nbytes"])
    t = torch.empty([0], device=metadata["device"], dtype=metadata["dtype"])
    t.set_(source=s, storage_offset=metadata["storage_offset"],
           size=metadata["size"], stride=metadata["stride"], )
    return t

def print_mem_stats(name):
    segments = torch.cuda.memory_snapshot()
    seg = []
    for segment in segments:
        if "segment_pool_id" in segment:
            tmp = ({"stream": segment["stream"]},
                   {"pool_id": segment["segment_pool_id"]},
                   {"block_num": len(segment["blocks"])},
                   {"activate_num": sum(int(blk["state"] == "active_allocated") for blk in segment["blocks"])},
                   {"total_size": segment["total_size"]},
                   {"allocated_size": segment["allocated_size"]},)
            seg.append(tmp)
    seg_str = "\n".join([str(seg_iter) for seg_iter in seg])
    seg_str = '\n' + seg_str
    print(f"{name}, snapshot: {seg_str}")

def cudagraphify(fn, inputs, pool, stream):
    torch.cuda.synchronize()
    gc.collect()
    torch.cuda.empty_cache()

    graph = torch.cuda.CUDAGraph()
    with torch.cuda.graph(graph, stream=stream, pool=pool):
        static_outputs = fn(*inputs)
    return graph, static_outputs


def foo(x, idx):
    r1 = x.expand([1, 2097152 // 8]).sqrt()
    r2 = x.expand([idx, 2097152]).clone()
    return (r1, r2)

# init
pool_id = torch.cuda.graph_pool_handle()
com_stream = torch.cuda.Stream()
com_device = torch.cuda.current_device()
inp = torch.tensor([7], device=com_device)

# record original state
with torch.cuda.stream(com_stream):
    g = torch.cuda.CUDAGraph()
    g.capture_begin(pool=pool_id)
    g.capture_end()
original_mem_state = torch._C._cuda_getCheckpointState(com_device, pool_id)

# start capture graph1
graph1, outputs1 = cudagraphify(foo, [inp, 1], pool=pool_id, stream=com_stream)
small_state = torch._C._cuda_getCheckpointState(com_device, pool_id)
print_mem_stats("\n-----------------after_small_state_run_g0_step0    ")
output1_metadata = [tensor_metadata(t) for t in outputs1]
outputs1 = None

# set to original state and capture graph2
torch._C._cuda_setCheckpointPoolState(com_device, original_mem_state, [], [])
print_mem_stats("\n-----------------after_set_origin_state     ")
graph2, outputs2 = cudagraphify(foo, [inp, 2], pool=pool_id, stream=com_stream)
biiig_state = torch._C._cuda_getCheckpointState(com_device, pool_id)
print_mem_stats("\n-----------------after_biiig_state_run_g1_step0    ")
output2_storage = [output.untyped_storage()._cdata for output in outputs2]

# set to original state and replay graph1
torch._C._cuda_setCheckpointPoolState(com_device, original_mem_state, output2_storage, [])
print_mem_stats("\n-----------------after_set_origin_state     ")
graph1.replay()
reconstructed_tensors1 = [reconstruct_from_tensor_metadata(metadata) for metadata in output1_metadata]
output1_new_storage = [output.untyped_storage()._cdata for output in reconstructed_tensors1]
torch._C._cuda_setCheckpointPoolState(com_device, small_state, [], output1_new_storage)
print_mem_stats("\n-----------------after_small_state_run_g0_step1    ")
```

After analyzing the error, the logic of the assertion seems to be: when allocing a new memory block, the last
block must be an unmapped 
nullptr block. Therefore, when setting to the recorded CheckpointState, the last block should also be unmapped
nullptr block.

Why does it fail only when enabling expandable_segments? It seems that after enabling expandable_segments, the 
reserved memory blocks 
will be merged. Therefore, when allocing a small block, releasing this small block, and then allocing a large
block again, the total reserved 
memory size increases. So, when setting to the state of a small memory block again, it triggers this assert 
error.

Could you please help check if these failed validations are always necessary? And how the current checkpoint 
memory management interface can solve the above problems when expandable_segments is enabled?  Thanks a lot.

### Versions

Collecting environment information...
PyTorch version: 2.6.0+cu124
Is debug build: False
CUDA used to build PyTorch: 12.4
ROCM used to build PyTorch: N/A

OS: Ubuntu 20.04.6 LTS (x86_64)
GCC version: (Ubuntu 9.4.0-1ubuntu1~20.04.3) 9.4.0
Clang version: Could not collect
CMake version: version 3.16.3
Libc version: glibc-2.31

Python version: 3.11.4 (main, Jul  5 2023, 13:45:01) [GCC 11.2.0] (64-bit runtime)
Python platform: Linux-5.15.0-72-generic-x86_64-with-glibc2.31
Is CUDA available: True
CUDA runtime version: 12.9.41
CUDA_MODULE_LOADING set to: LAZY
GPU models and configuration:
GPU 0: NVIDIA A100-SXM4-80GB
GPU 1: NVIDIA A100-SXM4-80GB
GPU 2: NVIDIA A100-SXM4-80GB
GPU 3: NVIDIA A100-SXM4-80GB
GPU 4: NVIDIA A100-SXM4-80GB
GPU 5: NVIDIA A100-SXM4-80GB
GPU 6: NVIDIA A100-SXM4-80GB
GPU 7: NVIDIA A100-SXM4-80GB

Nvidia driver version: 575.51.03
cuDNN version: Could not collect
HIP runtime version: N/A
MIOpen runtime version: N/A
Is XNNPACK available: True

\end{verbatim}
\end{tcolorbox}

\subsection{Issues and Comparative Solutions}

We present three representative cases that demonstrate divergent agent behaviors:

\textbf{CUDA Memory Checkpoint Assertion failure with Expandable Segments}

As shown above, PyTorch Issue \#161356 describes an error in PyTorch's CUDA graph checkpointing with expandable segments enabled. During a sequence that saves a small memory state, then a larger one, and restores the smaller state, PyTorch raises the error: ``Expected \small{\texttt{curr\_block->next == nullptr}} \normalsize to be true.'' Assertion failures occur in the allocator's checkpoint restoration logic, where terminal blocks in expandable segments are expected to have null next pointers.

Both frameworks identified the same underlying issue but proposed fundamentally different solutions. CCA viewed the assertions as overly restrictive and simply removed the problematic \small{\texttt{TORCH\_CHECK(curr\_block->next == nullptr)}} \normalsize assertions (--2 lines), while preserving other essential safety checks. CC viewed the assertions as important architectural guardrails and, instead of removing them, added logic (+7 lines) to explicitly set \small{\texttt{remaining->next = nullptr}} \normalsize during block splitting, effectively making expandable segments comply with the assertion. CCA favored minimal intervention, while CC pursued a more holistic solution. In this case, we note that the PyTorch team's eventual fix matched CCA's approach, providing human validation of CCA's principled engineering style.

\textbf{Excessive Memory Allocation and Deallocation during Llama-2 Training}

PyTorch Issue \#135837 highlights a memory allocation problem encountered during Llama-2 (70B) model Training. When GPU memory utilization approaches hardware limits, the allocator must decide whether to reclaim cached memory or retain it for performance optimization. On A100 GPUs (80GB), excessive memory deallocation and reallocation cycles occurred when reserved memory exceeded 70GB despite \small{\texttt{expandable\_segments=True}}, \normalsize resulting in significant Training slowdowns.

The issue was identified in PyTorch's CUDA allocator logic: even when the user explicitly enables expandable segments, the \small{\texttt{release\_cached\_blocks()}} \normalsize function continues to unmap expandable segments, causing unintended memory deallocation. This created a contradiction between user intent (maintaining expanded memory) and system behavior (aggressive memory reclamation). 

To solve this issue, CCA implemented a single guard clause (+6 lines) to fully disable memory reclamation when expandable segments are enabled, thus strictly adhering to the user's intent. In contrast, CC developed a more complex solution (+63 lines), dynamically measuring memory pressure and raising the reclamation threshold from 87.5\% to 95\%, while preserving emergency mechanisms above 95\% to prevent out-of-memory errors. Both solutions effectively addressed the immediate issue but reflected contrasting philosophies. CCA identified and minimally fixed the problematic behavior, while CC addressed the broader memory management challenge with a more sophisticated approach. 

\textbf{Neural Network Precision Regression during Weight Manipulation}

PyTorch Issue \#163072 involves a test failure where \small{\texttt{test\_partial\_flat\_weights}} \normalsize produces numerical accuracy errors on A100 and H100 GPUs, with relative differences of approximately 0.003 versus the expected tolerance of 1.3e-06. The test validates that partial flat weights (a memory optimization that stores select neural network layers in contiguous memory blocks) yield results equivalent to standard weight storage. The failure occurred during a sequence where LSTM weights were deleted, the module was moved to CUDA, and weights were manually restored. 

Both CCA and CC correctly identified the root cause as inappropriate test tolerances for GPU operations and  implemented the same adjustment: changing the tolerance parameters from exact equality to \small{\texttt{atol=1e-4,} \texttt{rtol=1e-4}} \normalsize (+1 line). This modification allowed the test to pass. 

However, CC did not stop there. While confirming that the adjusted test passes, CC encountered a benign warning about non-contiguous memory layout. In response, CC implemented additional changes (+20 lines), modifying \small{\texttt{setattr}} \normalsize in \small{\texttt{RNNBase}} \normalsize to automatically call \small{\texttt{flatten\_parameters()}} \normalsize whenever weights are manually assigned. Ultimately, CC addressed not only the immediate test failure but also the underlying warning, resulting in a more comprehensive but potentially over-engineered solution. 

\subsection{Architectural Analysis}

\begin{figure}[h]
  \centering
  \includegraphics[width=0.8\textwidth]{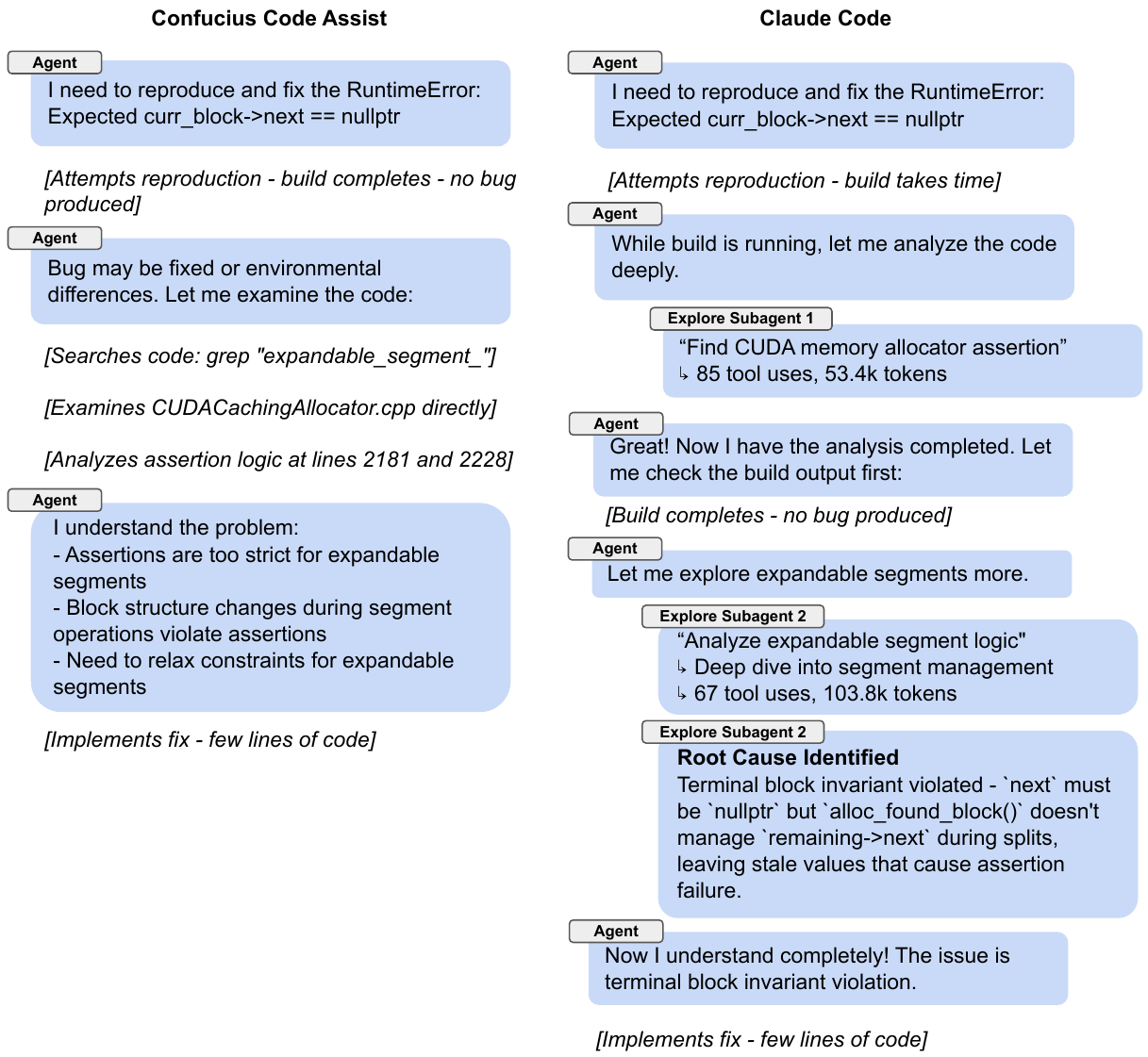}
  \caption{Simplified traces for CCA and CC on PyTorch issue \#161356}
  \label{fig:pytorch_traces}
\end{figure}

This section analyzes the problem-solving behavior of CCA versus CC on Issue \#161356. Figure~\ref{fig:pytorch_traces} shows simplified traces of the solutions of both agents. Both employed similar high-level strategies: (1) initial reproduction attempts, (2) code exploration when reproduction failed, (3) root cause identification through systematic analysis. 

Both agents also recognized the need for two directions of exploration: locating assertion failures in the CUDA memory allocator and understanding expandable segments logic. However, they perform these explorations differently: 
\begin{enumerate}
    \item \textbf{Single-Agent:} CCA performed explorations directly within the original context, maintaining awareness of the user's problem, system instructions, and previous observations. As such, CCA's explorations are subsequences of the overall reasoning chain.
    \item \textbf{Multi-Agent:} CC delegates investigations to separate, stateless subagents. As shown in Figure~\ref{fig:pytorch_traces},  CC even executes one of the subagents concurrently, while the main agent is running the reproduction script. These agents do not access the main agent's context, but they are initialized with a detailed prompt that emphasizes thoroughness (``use a thorough approach to find all relevant files''). 
\end{enumerate}

The architectural differences between CCA and CC significantly influenced solution characteristics. CCA's solution was simpler and more cautious, whereas CC was more ambitious and overengineered its solution. CC opted for this solution because it was the suggestion returned by the subagent, which was tasked with performing an exhaustive analysis. The subagent's mandate for thoroughness, combined with its lack of the original context, leads it to over-analyze the problem and provide a more complex solution than is necessary. The main agent, trusting the subagent's expertise, implements the solution despite preferring simpler solutions independently. This highlights a fundamental challenge in multi-agent systems. Generally, subagents separate concerns and allow the main agent to focus on its main task. However, our analysis suggests that for well-scoped debugging tasks, the benefits of delegation may be outweighed by the risk of context loss and derailment via inter-agent misalignment.

\section{Future Work}

Recent advances suggest that reinforcement learning (RL) can substantially enhance LLM-based software engineering agents beyond what is achievable with supervised fine-tuning alone. For example, SWE-RL~\citep{wei2025swe} demonstrates meaningful gains through end-to-end RL with verifiable rewards, while frameworks such as Agent Lightning~\citep{luo2025agentlightningtrainai} highlight a crucial architectural insight: by viewing agent execution as a Markov Decision Process, RL training can be decoupled from agent implementation via unified trajectory interfaces.
This decoupled perspective aligns naturally with the design philosophy behind CCA. The Agent Experience (AX) framework already structures an agent’s internal reasoning traces in a trajectory-friendly format, making them directly suitable for RL training. Moreover, CCA’s Meta-agent produces rich, fine-grained feedback signals from both tool extensions and environment interactions—signals that can be transformed into diverse reward functions for outcome-based, process-based, or hybrid supervision. Examples include rewards tied to note-taking quality, tool-use robustness, recovery behaviors, or the efficiency of multi-step exploration.
In addition, the extensibility of the Confucius Orchestrator provides a natural substrate for curriculum design in RL. Agent builders can introduce progressively richer toolsets, varied execution environments (e.g., shell, file editing, SQL databases), and increasingly complex tasks, enabling models to acquire generalizable agentic capabilities rather than overfitting to specific tool behaviors. This opens the door to RL-driven improvement not only of individual policies, but of the broader agent stack itself.

\clearpage

\end{document}